\PassOptionsToPackage{unicode}{hyperref}
\PassOptionsToPackage{hyphens}{url}
\PassOptionsToPackage{dvipsnames,svgnames,x11names}{xcolor}
\documentclass[
  letterpaper,
  DIV=11,
  numbers=noendperiod]{scrartcl}

\usepackage{amsmath,amssymb}
\usepackage{iftex}
\ifPDFTeX
  \usepackage[T1]{fontenc}
  \usepackage[utf8]{inputenc}
  \usepackage{textcomp} 
\else 
  \usepackage{unicode-math}
  \defaultfontfeatures{Scale=MatchLowercase}
  \defaultfontfeatures[\rmfamily]{Ligatures=TeX,Scale=1}
\fi
\usepackage{lmodern}
\ifPDFTeX\else  
\fi
\IfFileExists{upquote.sty}{\usepackage{upquote}}{}
\IfFileExists{microtype.sty}{
  \usepackage[]{microtype}
  \UseMicrotypeSet[protrusion]{basicmath} 
}{}
\makeatletter
\@ifundefined{KOMAClassName}{
  \IfFileExists{parskip.sty}{%
    \usepackage{parskip}
  }{
    \setlength{\parindent}{0pt}
    \setlength{\parskip}{6pt plus 2pt minus 1pt}}
}{
  \KOMAoptions{parskip=half}}
\makeatother
\usepackage{xcolor}
\ifx\paragraph\undefined\else
  \let\oldparagraph\paragraph
  \renewcommand{\paragraph}[1]{\oldparagraph{#1}\mbox{}}
\fi
\ifx\subparagraph\undefined\else
  \let\oldsubparagraph\subparagraph
  \renewcommand{\subparagraph}[1]{\oldsubparagraph{#1}\mbox{}}
\fi

\usepackage{color}
\usepackage{fancyvrb}

\DefineVerbatimEnvironment{Highlighting}{Verbatim}{commandchars=\\\{\}}
\usepackage{framed}
\definecolor{shadecolor}{RGB}{241,243,245}
\newenvironment{Shaded}{\begin{snugshade}}{\end{snugshade}}

\newcommand{\BuiltInTok}[1]{\textcolor[rgb]{0.00,0.23,0.31}{#1}}

\newcommand{\CommentTok}[1]{\textcolor[rgb]{0.37,0.37,0.37}{#1}}

\newcommand{\DataTypeTok}[1]{\textcolor[rgb]{0.68,0.00,0.00}{#1}}
\newcommand{\DecValTok}[1]{\textcolor[rgb]{0.68,0.00,0.00}{#1}}

\newcommand{\ErrorTok}[1]{\textcolor[rgb]{0.68,0.00,0.00}{#1}}

\newcommand{\FloatTok}[1]{\textcolor[rgb]{0.68,0.00,0.00}{#1}}
\newcommand{\FunctionTok}[1]{\textcolor[rgb]{0.28,0.35,0.67}{#1}}
\newcommand{\ImportTok}[1]{\textcolor[rgb]{0.00,0.46,0.62}{#1}}

\newcommand{\NormalTok}[1]{\textcolor[rgb]{0.00,0.23,0.31}{#1}}
\newcommand{\OperatorTok}[1]{\textcolor[rgb]{0.37,0.37,0.37}{#1}}
\newcommand{\OtherTok}[1]{\textcolor[rgb]{0.00,0.23,0.31}{#1}}

\newcommand{\StringTok}[1]{\textcolor[rgb]{0.13,0.47,0.30}{#1}}
\newcommand{\VariableTok}[1]{\textcolor[rgb]{0.07,0.07,0.07}{#1}}

\providecommand{\tightlist}{%
  \setlength{\itemsep}{0pt}\setlength{\parskip}{0pt}}\usepackage{longtable,booktabs,array}
\usepackage{calc} 
\usepackage{etoolbox}
\makeatletter
\patchcmd\longtable{\par}{\if@noskipsec\mbox{}\fi\par}{}{}
\makeatother
\IfFileExists{footnotehyper.sty}{\usepackage{footnotehyper}}{\usepackage{footnote}}
\makesavenoteenv{longtable}
\usepackage{graphicx}
\makeatletter
\def\maxwidth{\ifdim\Gin@nat@width>\linewidth\linewidth\else\Gin@nat@width\fi}
\def\maxheight{\ifdim\Gin@nat@height>\textheight\textheight\else\Gin@nat@height\fi}
\makeatother
\setkeys{Gin}{width=\maxwidth,height=\maxheight,keepaspectratio}
\makeatletter
\def\fps@figure{htbp}
\makeatother
\newlength{\cslhangindent}
\newlength{\csllabelwidth}
\newlength{\cslentryspacingunit} 
\newenvironment{CSLReferences}[2] 
 {
  \setlength{\parindent}{0pt}
  \ifodd #1
  \let\oldpar\par
  \def\par{\hangindent=\cslhangindent\oldpar}
  \fi
  \setlength{\parskip}{#2\cslentryspacingunit}
 }%
 {}
\usepackage{calc}

\usepackage{mdframed}
\usepackage{pseudo}
\usepackage{makeidx}
\usepackage{glossaries}
\usepackage{hyperref}
\makeglossaries
\newacronym{adwin}{ADWIN}{Adaptive Windowing}
\newacronym{alma}{ALMA}{Approximative Large-Margin-Algorithmus}
\newacronym{bip}{BIP}{Bruttoinlandsprodukt}
\newacronym{bml}{BML}{Batch Machine Learning}
\newacronym{bo}{BO}{Bayesian optimization}
\newacronym{cart}{CART}{Classification And Regression Tree}
\newacronym{cv}{CV}{Cross Validation}
\newacronym{cvfdt}{CVFDT}{Concept-adapting Very Fast Decision Tree}
\newacronym{dace}{DACE}{Design and Analysis of Computer Experiments}
\newacronym{ddm}{DDM}{Drift Detection Method}
\newacronym{dl}{DL}{Deep Learning}
\newacronym{doe}{DOE}{Design of Experiments}
\newacronym{efdt}{EFDT}{Extremely Fast Decision Tree}
\newacronym{gbrt}{gbrt}{Gradient Boosting Regression Tree}
\newacronym{gcd}{GCD}{Greatest Common Divisor}
\newacronym{gra}{GRA}{Global Recurring Abrupt}
\newacronym{hat}{HAT}{Hoeffding Adaptive Tree}
\newacronym{hatc}{HATC}{Hoeffding Adaptive Tree Classifier}
\newacronym{hatr}{HATR}{Hoeffding Adaptive Tree Regressor}
\newacronym{hpt}{HPT}{Hyperparameter Tuning}
\newacronym{ht}{HT}{Hoeffding Tree}
\newacronym{htc}{HTC}{Hoeffding Tree Classifier}
\newacronym{htr}{HTR}{Hoeffding Tree Regressor}
\newacronym{ki}{KI}{Künstliche Intelligenz}
\newacronym{kpi}{KPI}{Key Performance Indicator}
\newacronym{mae}{MAE}{Mean Absolute Error}
\newacronym{ml}{ML}{Machine Learning}
\newacronym{moa}{MOA}{Massive Online Analysis}
\newacronym{mse}{MSE}{Mean Squared Error}
\newacronym{nn}{NN}{Neural Network}
\newacronym{oml}{OML}{Online Machine Learning}
\newacronym{pa}{PA}{Passive-Aggressive}
\newacronym{pca}{PCA}{Principal Component Analysis}
\newacronym{raytune}{ray[tune]}{Ray Tune}
\newacronym{rf}{RF}{Random Forest}
\newacronym{river}{river}{River: Online machine learning in Python}
\newacronym{rmoa}{RMOA}{Massive Online Analysis in R}
\newacronym{rocauc}{ROC AUC}{AUC (Area Under The Curve) ROC (Receiver Operating Characteristics)}
\newacronym{sea}{SEA}{SEA synthetic dataset}
\newacronym{sklearn}{sklearn}{scikit-learn}
\newacronym{smbo}{SMBO}{Surrogate Model Based Optimization}
\newacronym{smote}{SMOTE}{Synthetic Minority Oversampling Technique}
\newacronym{spo}{SPO}{Sequential Parameter Optimization}
\newacronym{spot}{SPOT}{Sequential Parameter Optimization Toolbox}
\newacronym{spotpython}{spotPython}{Sequential Parameter Optimization Toolbox for Python}
\newacronym{spotriver}{spotRiver}{Sequential Parameter Optimization Toolbox for River}
\newacronym{sgd}{SGD}{Stochastic Gradient Descent}
\newacronym{svm}{SVM}{Support Vector Machine}
\newacronym{vfdt}{VFDT}{Very Fast Decision Tree}
\KOMAoption{captions}{tableheading}
\makeatletter
\@ifpackageloaded{tcolorbox}{}{\usepackage[skins,breakable]{tcolorbox}}
\@ifpackageloaded{fontawesome5}{}{\usepackage{fontawesome5}}
\definecolor{quarto-callout-color}{HTML}{909090}
\definecolor{quarto-callout-note-color}{HTML}{0758E5}
\definecolor{quarto-callout-important-color}{HTML}{CC1914}
\definecolor{quarto-callout-warning-color}{HTML}{EB9113}
\definecolor{quarto-callout-tip-color}{HTML}{00A047}
\definecolor{quarto-callout-caution-color}{HTML}{FC5300}
\definecolor{quarto-callout-color-frame}{HTML}{acacac}
\definecolor{quarto-callout-note-color-frame}{HTML}{4582ec}
\definecolor{quarto-callout-important-color-frame}{HTML}{d9534f}
\definecolor{quarto-callout-warning-color-frame}{HTML}{f0ad4e}
\definecolor{quarto-callout-tip-color-frame}{HTML}{02b875}
\definecolor{quarto-callout-caution-color-frame}{HTML}{fd7e14}
\makeatother
\makeatletter
\@ifpackageloaded{caption}{}{\usepackage{caption}}
\AtBeginDocument{%
\ifdefined\contentsname
  \renewcommand*\contentsname{Table of contents}
\else
  \newcommand\contentsname{Table of contents}
\fi
\ifdefined\listfigurename
  \renewcommand*\listfigurename{List of Figures}
\else
  \newcommand\listfigurename{List of Figures}
\fi
\ifdefined\listtablename
  \renewcommand*\listtablename{List of Tables}
\else
  \newcommand\listtablename{List of Tables}
\fi
\ifdefined\figurename
  \renewcommand*\figurename{Figure}
\else
  \newcommand\figurename{Figure}
\fi
\ifdefined\tablename
  \renewcommand*\tablename{Table}
\else
  \newcommand\tablename{Table}
\fi
}
\@ifpackageloaded{float}{}{\usepackage{float}}
\floatstyle{ruled}
\@ifundefined{c@chapter}{\newfloat{codelisting}{h}{lop}}{\newfloat{codelisting}{h}{lop}[chapter]}
\floatname{codelisting}{Listing}

\makeatother
\makeatletter
\@ifpackageloaded{caption}{}{\usepackage{caption}}
\@ifpackageloaded{subcaption}{}{\usepackage{subcaption}}
\makeatother
\makeatletter
\@ifpackageloaded{tcolorbox}{}{\usepackage[skins,breakable]{tcolorbox}}
\makeatother
\makeatletter
\@ifundefined{shadecolor}{\definecolor{shadecolor}{rgb}{.97, .97, .97}}
\makeatother
\ifLuaTeX
  \usepackage{selnolig}  
\fi
\IfFileExists{bookmark.sty}{\usepackage{bookmark}}{\usepackage{hyperref}}
\IfFileExists{xurl.sty}{\usepackage{xurl}}{} 
\hypersetup{
  pdftitle={PyTorch Hyperparameter Tuning --- A Tutorial for spotPython},
  pdfauthor={Thomas Bartz-Beielstein},
  pdfkeywords={hyperparameter tuning, hyperparameter
optimization, spotPython, PyTorch, CIFAR10, optimization, deep
learning, machine learning},
  colorlinks=true,
  linkcolor={blue},
  filecolor={Maroon},
  citecolor={Blue},
  urlcolor={Blue},
  pdfcreator={LaTeX via pandoc}}

\title{PyTorch Hyperparameter Tuning --- A Tutorial for spotPython}
\usepackage{etoolbox}
\makeatletter
\providecommand{\subtitle}[1]{
  \apptocmd{\@title}{\par {\large #1 \par}}{}{}
}
\makeatother
\subtitle{Version 0.2.15}
\author{Thomas Bartz-Beielstein\\
bartzbeielstein@gmail.com\\
\url{https://orcid.org/0000-0002-5938-5158}}
\date{June, 7th 2023}
\begin{document}
\maketitle
\begin{abstract}
The goal of hyperparameter tuning (or hyperparameter optimization) is to optimize the hyperparameters to improve the performance of the machine or deep learning model. spotPython (``Sequential Parameter Optimization Toolbox in Python'') is the Python version of the well-known hyperparameter tuner SPOT, which has been developed in the R programming environment for statistical analysis for over a decade. PyTorch is an optimized tensor library for deep learning using GPUs and CPUs. This document shows how to integrate the spotPython hyperparameter tuner into the PyTorch training workflow.  As an example, the results of the CIFAR10 image classifier are used. In addition to an introduction to spotPython, this tutorial also includes a brief comparison with Ray Tune, a Python library for running experiments and tuning hyperparameters. This comparison is based on the PyTorch hyperparameter tuning tutorial. The advantages and disadvantages of both approaches are discussed. We show that spotPython achieves similar or even better results while being more flexible and transparent than Ray Tune.
\end{abstract}
\hypertarget{sec-hyperparameter-tuning}{%
\section{Hyperparameter Tuning}\label{sec-hyperparameter-tuning}}

Hyperparameter tuning is an important, but often difficult and
computationally intensive task. Changing the architecture of a neural
network or the learning rate of an optimizer can have a significant
impact on the performance.

The goal of hyperparameter tuning is to optimize the hyperparameters in
a way that improves the performance of the machine learning or deep
learning model. The simplest, but also most computationally expensive,
approach uses manual search (or trial-and-error (Meignan et al. 2015)).
Commonly encountered is simple random search, i.e., random and repeated
selection of hyperparameters for evaluation, and lattice search (``grid
search''). In addition, methods that perform directed search and other
model-free algorithms, i.e., algorithms that do not explicitly rely on a
model, e.g., evolution strategies (Bartz-Beielstein et al. 2014) or
pattern search (Lewis, Torczon, and Trosset 2000) play an important
role. Also, ``hyperband'', i.e., a multi-armed bandit strategy that
dynamically allocates resources to a set of random configurations and
uses successive bisections to stop configurations with poor performance
(Li et al. 2016), is very common in hyperparameter tuning. The most
sophisticated and efficient approaches are the Bayesian optimization and
surrogate model based optimization methods, which are based on the
optimization of cost functions determined by simulations or experiments.

We consider below a surrogate model based optimization-based
hyperparameter tuning approach based on the Python version of the SPOT
(``Sequential Parameter Optimization Toolbox'') (Bartz-Beielstein,
Lasarczyk, and Preuss 2005), which is suitable for situations where only
limited resources are available. This may be due to limited availability
and cost of hardware, or due to the fact that confidential data may only
be processed locally, e.g., due to legal requirements. Furthermore, in
our approach, the understanding of algorithms is seen as a key tool for
enabling transparency and explainability. This can be enabled, for
example, by quantifying the contribution of machine learning and deep
learning components (nodes, layers, split decisions, activation
functions, etc.). Understanding the importance of hyperparameters and
the interactions between multiple hyperparameters plays a major role in
the interpretability and explainability of machine learning models. SPOT
provides statistical tools for understanding hyperparameters and their
interactions. Last but not least, it should be noted that the SPOT
software code is available in the open source \texttt{spotPython}
package on github\footnote{\url{https://github.com/sequential-parameter-optimization}},
allowing replicability of the results. This tutorial descries the Python
variant of SPOT, which is called \texttt{spotPython}. The R
implementation is described in Bartz et al. (2022). SPOT is an
established open source software that has been maintained for more than
15 years (Bartz-Beielstein, Lasarczyk, and Preuss 2005) (Bartz et al.
2022).

This tutorial is structured as follows. The concept of the
hyperparameter tuning software \texttt{spotPython} is described in
Section~\ref{sec-spot}.
Section~\ref{sec-hyperparameter-tuning-for-pytorch} describes the
execution of the example from the tutorial ``Hyperparameter Tuning with
Ray Tune'' (PyTorch 2023a). It describes the integration of
\texttt{spotPython} into the \texttt{PyTorch} training workflow in
detail and presents the results. Finally, Section~\ref{sec-summary}
presents a summary and an outlook.

\begin{tcolorbox}[enhanced jigsaw, colbacktitle=quarto-callout-note-color!10!white, coltitle=black, opacitybacktitle=0.6, title=\textcolor{quarto-callout-note-color}{\faInfo}\hspace{0.5em}{Note}, bottomrule=.15mm, arc=.35mm, leftrule=.75mm, breakable, colframe=quarto-callout-note-color-frame, colback=white, left=2mm, toptitle=1mm, bottomtitle=1mm, rightrule=.15mm, titlerule=0mm, toprule=.15mm, opacityback=0]

The corresponding \texttt{.ipynb} notebook (Bartz-Beielstein 2023) is
updated regularly and reflects updates and changes in the
\texttt{spotPython} package. It can be downloaded from
\url{https://github.com/sequential-parameter-optimization/spotPython/blob/main/notebooks/14_spot_ray_hpt_torch_cifar10.ipynb}.

\end{tcolorbox}

\hypertarget{sec-spot}{%
\section{The Hyperparameter Tuning Software SPOT}\label{sec-spot}}

Surrogate model based optimization methods are common approaches in
simulation and optimization. SPOT was developed because there is a great
need for sound statistical analysis of simulation and optimization
algorithms. SPOT includes methods for tuning based on classical
regression and analysis of variance techniques. It presents tree-based
models such as classification and regression trees and random forests as
well as Bayesian optimization (Gaussian process models, also known as
Kriging). Combinations of different meta-modeling approaches are
possible. SPOT comes with a sophisticated surrogate model based
optimization method, that can handle discrete and continuous inputs.
Furthermore, any model implemented in \texttt{scikit-learn} can be used
out-of-the-box as a surrogate in \texttt{spotPython}.

SPOT implements key techniques such as exploratory fitness landscape
analysis and sensitivity analysis. It can be used to understand the
performance of various algorithms, while simultaneously giving insights
into their algorithmic behavior. In addition, SPOT can be used as an
optimizer and for automatic and interactive tuning. Details on SPOT and
its use in practice are given by Bartz et al. (2022).

A typical hyperparameter tuning process with \texttt{spotPython}
consists of the following steps:

\begin{enumerate}
\def\labelenumi{\arabic{enumi}.}
\tightlist
\item
  Loading the data (training and test datasets), see
  Section~\ref{sec-data-loading}.
\item
  Specification of the preprocessing model, see
  Section~\ref{sec-specification-of-preprocessing-model}. This model is
  called \texttt{prep\_model} (``preparation'' or pre-processing). The
  information required for the hyperparameter tuning is stored in the
  dictionary \texttt{fun\_control}. Thus, the information needed for the
  execution of the hyperparameter tuning is available in a readable
  form.
\item
  Selection of the machine learning or deep learning model to be tuned,
  see Section~\ref{sec-selection-of-the-algorithm}. This is called the
  \texttt{core\_model}. Once the \texttt{core\_model} is defined, then
  the associated hyperparameters are stored in the \texttt{fun\_control}
  dictionary. First, the hyperparameters of the \texttt{core\_model} are
  initialized with the default values of the \texttt{core\_model}. As
  default values we use the default values contained in the
  \texttt{spotPython} package for the algorithms of the \texttt{torch}
  package.
\item
  Modification of the default values for the hyperparameters used in
  \texttt{core\_model}, see
  Section~\ref{sec-modification-of-default-values}. This step is
  optional.

  \begin{enumerate}
  \def\labelenumii{\arabic{enumii}.}
  \tightlist
  \item
    numeric parameters are modified by changing the bounds.
  \item
    categorical parameters are modified by changing the categories
    (``levels'').
  \end{enumerate}
\item
  Selection of target function (loss function) for the optimizer, see
  Section~\ref{sec-selection-of-target-function}.
\item
  Calling SPOT with the corresponding parameters, see
  Section~\ref{sec-call-the-hyperparameter-tuner}. The results are
  stored in a dictionary and are available for further analysis.
\item
  Presentation, visualization and interpretation of the results, see
  Section~\ref{sec-results-tuning}.
\end{enumerate}

\hypertarget{sec-hyperparameter-tuning-for-pytorch}{%
\section{\texorpdfstring{Hyperparameter Tuning for PyTorch With
\texttt{spotPython}}{Hyperparameter Tuning for PyTorch With spotPython}}\label{sec-hyperparameter-tuning-for-pytorch}}

In this tutorial, we will show how \texttt{spotPython} can be integrated
into the \texttt{PyTorch} training workflow. It is based on the tutorial
``Hyperparameter Tuning with Ray Tune'' from the \texttt{PyTorch}
documentation (PyTorch 2023a), which is an extension of the tutorial
``Training a Classifier'' (PyTorch 2023b) for training a CIFAR10 image
classifier.

This document refers to the following software versions:

\begin{itemize}
\tightlist
\item
  \texttt{python}: 3.10.10
\item
  \texttt{torch}: 2.0.1
\item
  \texttt{torchvision}: 0.15.0
\item
  \texttt{spotPython}: 0.2.15
\end{itemize}

\texttt{spotPython} can be installed via pip\footnote{Alternatively, the
  source code can be downloaded from gitHub:
  \url{https://github.com/sequential-parameter-optimization/spotPython}.}.

\begin{Shaded}
\begin{Highlighting}[]
\NormalTok{!pip install spotPython}
\end{Highlighting}
\end{Shaded}

Results that refer to the \texttt{Ray\ Tune} package are taken from
\url{https://PyTorch.org/tutorials/beginner/hyperparameter_tuning_tutorial.html}\footnote{We
  were not able to install \texttt{Ray\ Tune} on our system. Therefore,
  we used the results from the \texttt{PyTorch} tutorial.}.

\hypertarget{sec-setup}{%
\subsection{Setup}\label{sec-setup}}

Before we consider the detailed experimental setup, we select the
parameters that affect run time, initial design size and the device that
is used.

\begin{Shaded}
\begin{Highlighting}[]
\NormalTok{MAX\_TIME }\OperatorTok{=} \DecValTok{60}
\NormalTok{INIT\_SIZE }\OperatorTok{=} \DecValTok{20}
\NormalTok{DEVICE }\OperatorTok{=} \StringTok{"cpu"} \CommentTok{\# "cuda:0"}
\end{Highlighting}
\end{Shaded}

\hypertarget{initialization-of-the-fun_control-dictionary}{%
\subsection{\texorpdfstring{Initialization of the \texttt{fun\_control}
Dictionary}{Initialization of the fun\_control Dictionary}}\label{initialization-of-the-fun_control-dictionary}}

\texttt{spotPython} uses a Python dictionary for storing the information
required for the hyperparameter tuning process. This dictionary is
called \texttt{fun\_control} and is initialized with the function
\texttt{fun\_control\_init}. The function \texttt{fun\_control\_init}
returns a skeleton dictionary. The dictionary is filled with the
required information for the hyperparameter tuning process. It stores
the hyperparameter tuning settings, e.g., the deep learning network
architecture that should be tuned, the classification (or regression)
problem, and the data that is used for the tuning. The dictionary is
used as an input for the SPOT function.

\begin{Shaded}
\begin{Highlighting}[]
\ImportTok{from}\NormalTok{ spotPython.utils.init }\ImportTok{import}\NormalTok{ fun\_control\_init}
\NormalTok{fun\_control }\OperatorTok{=}\NormalTok{ fun\_control\_init(task}\OperatorTok{=}\StringTok{"classification"}\NormalTok{,}
\NormalTok{    tensorboard\_path}\OperatorTok{=}\StringTok{"runs/14\_spot\_ray\_hpt\_torch\_cifar10"}\NormalTok{)}
\end{Highlighting}
\end{Shaded}

\hypertarget{sec-data-loading}{%
\subsection{Data Loading}\label{sec-data-loading}}

The data loading process is implemented in the same manner as described
in the Section ``Data loaders'' in PyTorch (2023a). The data loaders are
wrapped into the function \texttt{load\_data\_cifar10} which is
identical to the function \texttt{load\_data} in PyTorch (2023a). A
global data directory is used, which allows sharing the data directory
between different trials. The method \texttt{load\_data\_cifar10} is
part of the \texttt{spotPython} package and can be imported from
\texttt{spotPython.data.torchdata}.

In the following step, the test and train data are added to the
dictionary \texttt{fun\_control}.

\begin{Shaded}
\begin{Highlighting}[]
\ImportTok{from}\NormalTok{ spotPython.data.torchdata }\ImportTok{import}\NormalTok{ load\_data\_cifar10}
\NormalTok{train, test }\OperatorTok{=}\NormalTok{ load\_data\_cifar10()}
\NormalTok{n\_samples }\OperatorTok{=} \BuiltInTok{len}\NormalTok{(train)}
\CommentTok{\# add the dataset to the fun\_control}
\NormalTok{fun\_control.update(\{}
    \StringTok{"train"}\NormalTok{: train,}
    \StringTok{"test"}\NormalTok{: test,}
    \StringTok{"n\_samples"}\NormalTok{: n\_samples\})}
\end{Highlighting}
\end{Shaded}

\hypertarget{sec-specification-of-preprocessing-model}{%
\subsection{Specification of the Preprocessing
Model}\label{sec-specification-of-preprocessing-model}}

After the training and test data are specified and added to the
\texttt{fun\_control} dictionary, \texttt{spotPython} allows the
specification of a data preprocessing pipeline, e.g., for the scaling of
the data or for the one-hot encoding of categorical variables. The
preprocessing model is called \texttt{prep\_model} (``preparation'' or
pre-processing) and includes steps that are not subject to the
hyperparameter tuning process. The preprocessing model is specified in
the \texttt{fun\_control} dictionary. The preprocessing model can be
implemented as a \texttt{sklearn} pipeline. The following code shows a
typical preprocessing pipeline:

\begin{Shaded}
\begin{Highlighting}[]
\NormalTok{categorical\_columns = ["cities", "colors"]}
\NormalTok{one\_hot\_encoder = OneHotEncoder(handle\_unknown="ignore",}
\NormalTok{                                    sparse\_output=False)}
\NormalTok{prep\_model = ColumnTransformer(}
\NormalTok{        transformers=[}
\NormalTok{             ("categorical", one\_hot\_encoder, categorical\_columns),}
\NormalTok{         ],}
\NormalTok{         remainder=StandardScaler(),}
\NormalTok{     )}
\end{Highlighting}
\end{Shaded}

Because the Ray Tune (\texttt{ray{[}tune{]}}) hyperparameter tuning as
described in PyTorch (2023a) does not use a preprocessing model, the
preprocessing model is set to \texttt{None} here.

\begin{Shaded}
\begin{Highlighting}[]
\NormalTok{prep\_model }\OperatorTok{=} \VariableTok{None}
\NormalTok{fun\_control.update(\{}\StringTok{"prep\_model"}\NormalTok{: prep\_model\})}
\end{Highlighting}
\end{Shaded}

\hypertarget{sec-selection-of-the-algorithm}{%
\subsection{\texorpdfstring{Select \texttt{algorithm} and
\texttt{core\_model\_hyper\_dict}}{Select algorithm and core\_model\_hyper\_dict}}\label{sec-selection-of-the-algorithm}}

The same neural network model as implemented in the section
``Configurable neural network'' of the \texttt{PyTorch} tutorial
(PyTorch 2023a) is used here. We will show the implementation from
PyTorch (2023a) in Section~\ref{sec-implementation-with-raytune} first,
before the extended implementation with \texttt{spotPython} is shown in
Section~\ref{sec-implementation-with-spotpython}.

\hypertarget{sec-implementation-with-raytune}{%
\subsubsection{Implementing a Configurable Neural Network With Ray
Tune}\label{sec-implementation-with-raytune}}

We used the same hyperparameters that are implemented as configurable in
the \texttt{PyTorch} tutorial. We specify the layer sizes, namely
\texttt{l1} and \texttt{l2}, of the fully connected layers:

\begin{Shaded}
\begin{Highlighting}[]
\NormalTok{class Net(nn.Module):}
\NormalTok{    def \_\_init\_\_(self, l1=120, l2=84):}
\NormalTok{        super(Net, self).\_\_init\_\_()}
\NormalTok{        self.conv1 = nn.Conv2d(3, 6, 5)}
\NormalTok{        self.pool = nn.MaxPool2d(2, 2)}
\NormalTok{        self.conv2 = nn.Conv2d(6, 16, 5)}
\NormalTok{        self.fc1 = nn.Linear(16 * 5 * 5, l1)}
\NormalTok{        self.fc2 = nn.Linear(l1, l2)}
\NormalTok{        self.fc3 = nn.Linear(l2, 10)}

\NormalTok{    def forward(self, x):}
\NormalTok{        x = self.pool(F.relu(self.conv1(x)))}
\NormalTok{        x = self.pool(F.relu(self.conv2(x)))}
\NormalTok{        x = x.view({-}1, 16 * 5 * 5)}
\NormalTok{        x = F.relu(self.fc1(x))}
\NormalTok{        x = F.relu(self.fc2(x))}
\NormalTok{        x = self.fc3(x)}
\NormalTok{        return x}
\end{Highlighting}
\end{Shaded}

The learning rate, i.e., \texttt{lr}, of the optimizer is made
configurable, too:

\begin{Shaded}
\begin{Highlighting}[]
\NormalTok{optimizer = optim.SGD(net.parameters(), lr=config["lr"], momentum=0.9)}
\end{Highlighting}
\end{Shaded}

\hypertarget{sec-implementation-with-spotpython}{%
\subsubsection{Implementing a Configurable Neural Network With
spotPython}\label{sec-implementation-with-spotpython}}

\texttt{spotPython} implements a class which is similar to the class
described in the \texttt{PyTorch} tutorial. The class is called
\texttt{Net\_CIFAR10} and is implemented in the file
\texttt{netcifar10.py}.

\begin{Shaded}
\begin{Highlighting}[]
\NormalTok{from torch import nn}
\NormalTok{import torch.nn.functional as F}
\NormalTok{import spotPython.torch.netcore as netcore}


\NormalTok{class Net\_CIFAR10(netcore.Net\_Core):}
\NormalTok{    def \_\_init\_\_(self, l1, l2, lr\_mult, batch\_size, epochs, k\_folds, patience,}
\NormalTok{    optimizer, sgd\_momentum):}
\NormalTok{        super(Net\_CIFAR10, self).\_\_init\_\_(}
\NormalTok{            lr\_mult=lr\_mult,}
\NormalTok{            batch\_size=batch\_size,}
\NormalTok{            epochs=epochs,}
\NormalTok{            k\_folds=k\_folds,}
\NormalTok{            patience=patience,}
\NormalTok{            optimizer=optimizer,}
\NormalTok{            sgd\_momentum=sgd\_momentum,}
\NormalTok{        )}
\NormalTok{        self.conv1 = nn.Conv2d(3, 6, 5)}
\NormalTok{        self.pool = nn.MaxPool2d(2, 2)}
\NormalTok{        self.conv2 = nn.Conv2d(6, 16, 5)}
\NormalTok{        self.fc1 = nn.Linear(16 * 5 * 5, l1)}
\NormalTok{        self.fc2 = nn.Linear(l1, l2)}
\NormalTok{        self.fc3 = nn.Linear(l2, 10)}

\NormalTok{    def forward(self, x):}
\NormalTok{        x = self.pool(F.relu(self.conv1(x)))}
\NormalTok{        x = self.pool(F.relu(self.conv2(x)))}
\NormalTok{        x = x.view({-}1, 16 * 5 * 5)}
\NormalTok{        x = F.relu(self.fc1(x))}
\NormalTok{        x = F.relu(self.fc2(x))}
\NormalTok{        x = self.fc3(x)}
\NormalTok{        return x}
\end{Highlighting}
\end{Shaded}

\hypertarget{the-net_core-class}{%
\paragraph{\texorpdfstring{The \texttt{Net\_Core}
class}{The Net\_Core class}}\label{the-net_core-class}}

\texttt{Net\_CIFAR10} inherits from the class \texttt{Net\_Core} which
is implemented in the file \texttt{netcore.py}. It implements the
additional attributes that are common to all neural network models. The
\texttt{Net\_Core} class is implemented in the file \texttt{netcore.py}.
It implements hyperparameters as attributes, that are not used by the
\texttt{core\_model}, e.g.:

\begin{itemize}
\tightlist
\item
  optimizer (\texttt{optimizer}),
\item
  learning rate (\texttt{lr}),
\item
  batch size (\texttt{batch\_size}),
\item
  epochs (\texttt{epochs}),
\item
  k\_folds (\texttt{k\_folds}), and
\item
  early stopping criterion ``patience'' (\texttt{patience}).
\end{itemize}

Users can add further attributes to the class. The class
\texttt{Net\_Core} is shown below.

\begin{Shaded}
\begin{Highlighting}[]
\NormalTok{from torch import nn}


\NormalTok{class Net\_Core(nn.Module):}
\NormalTok{    def \_\_init\_\_(self, lr\_mult, batch\_size, epochs, k\_folds, patience,}
\NormalTok{        optimizer, sgd\_momentum):}
\NormalTok{        super(Net\_Core, self).\_\_init\_\_()}
\NormalTok{        self.lr\_mult = lr\_mult}
\NormalTok{        self.batch\_size = batch\_size}
\NormalTok{        self.epochs = epochs}
\NormalTok{        self.k\_folds = k\_folds}
\NormalTok{        self.patience = patience}
\NormalTok{        self.optimizer = optimizer}
\NormalTok{        self.sgd\_momentum = sgd\_momentum}
\end{Highlighting}
\end{Shaded}

\hypertarget{sec-comparison}{%
\subsubsection{Comparison of the Approach Described in the PyTorch
Tutorial With spotPython}\label{sec-comparison}}

Comparing the class \texttt{Net} from the \texttt{PyTorch} tutorial and
the class \texttt{Net\_CIFAR10} from \texttt{spotPython}, we see that
the class \texttt{Net\_CIFAR10} has additional attributes and does not
inherit from \texttt{nn} directly. It adds an additional class,
\texttt{Net\_core}, that takes care of additional attributes that are
common to all neural network models, e.g., the learning rate multiplier
\texttt{lr\_mult} or the batch size \texttt{batch\_size}.

\texttt{spotPython}'s \texttt{core\_model} implements an instance of the
\texttt{Net\_CIFAR10} class. In addition to the basic neural network
model, the \texttt{core\_model} can use these additional attributes.
\texttt{spotPython} provides methods for handling these additional
attributes to guarantee 100\% compatibility with the \texttt{PyTorch}
classes. The method \texttt{add\_core\_model\_to\_fun\_control} adds the
hyperparameters and additional attributes to the \texttt{fun\_control}
dictionary. The method is shown below.

\begin{Shaded}
\begin{Highlighting}[]
\ImportTok{from}\NormalTok{ spotPython.torch.netcifar10 }\ImportTok{import}\NormalTok{ Net\_CIFAR10}
\ImportTok{from}\NormalTok{ spotPython.data.torch\_hyper\_dict }\ImportTok{import}\NormalTok{ TorchHyperDict}
\ImportTok{from}\NormalTok{ spotPython.hyperparameters.values }\ImportTok{import}\NormalTok{ add\_core\_model\_to\_fun\_control}
\NormalTok{core\_model }\OperatorTok{=}\NormalTok{ Net\_CIFAR10}
\NormalTok{fun\_control }\OperatorTok{=}\NormalTok{ add\_core\_model\_to\_fun\_control(core\_model}\OperatorTok{=}\NormalTok{core\_model,}
\NormalTok{                              fun\_control}\OperatorTok{=}\NormalTok{fun\_control,}
\NormalTok{                              hyper\_dict}\OperatorTok{=}\NormalTok{TorchHyperDict,}
\NormalTok{                              filename}\OperatorTok{=}\VariableTok{None}\NormalTok{)}
\end{Highlighting}
\end{Shaded}

\hypertarget{sec-search-space}{%
\subsection{The Search Space}\label{sec-search-space}}

In Section~\ref{sec-configuring-the-search-space-with-ray-tune}, we
first describe how to configure the search space with
\texttt{ray{[}tune{]}} (as shown in PyTorch (2023a)) and then how to
configure the search space with \texttt{spotPython} in
Section~\ref{sec-configuring-the-search-space-with-spotpython}.

\hypertarget{sec-configuring-the-search-space-with-ray-tune}{%
\subsubsection{Configuring the Search Space With Ray
Tune}\label{sec-configuring-the-search-space-with-ray-tune}}

Ray Tune's search space can be configured as follows (PyTorch 2023a):

\begin{Shaded}
\begin{Highlighting}[]
\NormalTok{config = \{}
\NormalTok{    "l1": tune.sample\_from(lambda \_: 2**np.random.randint(2, 9)),}
\NormalTok{    "l2": tune.sample\_from(lambda \_: 2**np.random.randint(2, 9)),}
\NormalTok{    "lr": tune.loguniform(1e{-}4, 1e{-}1),}
\NormalTok{    "batch\_size": tune.choice([2, 4, 8, 16])}
\NormalTok{\}}
\end{Highlighting}
\end{Shaded}

The \texttt{tune.sample\_from()} function enables the user to define
sample methods to obtain hyperparameters. In this example, the
\texttt{l1} and \texttt{l2} parameters should be powers of 2 between 4
and 256, so either 4, 8, 16, 32, 64, 128, or 256. The \texttt{lr}
(learning rate) should be uniformly sampled between 0.0001 and 0.1.
Lastly, the batch size is a choice between 2, 4, 8, and 16.

At each trial, \texttt{ray{[}tune{]}} will randomly sample a combination
of parameters from these search spaces. It will then train a number of
models in parallel and find the best performing one among these.
\texttt{ray{[}tune{]}} uses the \texttt{ASHAScheduler} which will
terminate bad performing trials early.

\hypertarget{sec-configuring-the-search-space-with-spotpython}{%
\subsubsection{Configuring the Search Space With
spotPython}\label{sec-configuring-the-search-space-with-spotpython}}

\hypertarget{the-hyper_dict-hyperparameters-for-the-selected-algorithm}{%
\paragraph{\texorpdfstring{The \texttt{hyper\_dict} Hyperparameters for
the Selected
Algorithm}{The hyper\_dict Hyperparameters for the Selected Algorithm}}\label{the-hyper_dict-hyperparameters-for-the-selected-algorithm}}

\texttt{spotPython} uses \texttt{JSON} files for the specification of
the hyperparameters. Users can specify their individual \texttt{JSON}
files, or they can use the \texttt{JSON} files provided by
\texttt{spotPython}. The \texttt{JSON} file for the \texttt{core\_model}
is called \texttt{torch\_hyper\_dict.json}.

In contrast to \texttt{ray{[}tune{]}}, \texttt{spotPython} can handle
numerical, boolean, and categorical hyperparameters. They can be
specified in the \texttt{JSON} file in a similar way as the numerical
hyperparameters as shown below. Each entry in the \texttt{JSON} file
represents one hyperparameter with the following structure:
\texttt{type}, \texttt{default}, \texttt{transform}, \texttt{lower}, and
\texttt{upper}.

\begin{Shaded}
\begin{Highlighting}[]
\ErrorTok{"factor\_hyperparameter":} \FunctionTok{\{}
    \DataTypeTok{"levels"}\FunctionTok{:} \OtherTok{[}\StringTok{"A"}\OtherTok{,} \StringTok{"B"}\OtherTok{,} \StringTok{"C"}\OtherTok{]}\FunctionTok{,}
    \DataTypeTok{"type"}\FunctionTok{:} \StringTok{"factor"}\FunctionTok{,}
    \DataTypeTok{"default"}\FunctionTok{:} \StringTok{"B"}\FunctionTok{,}
    \DataTypeTok{"transform"}\FunctionTok{:} \StringTok{"None"}\FunctionTok{,}
    \DataTypeTok{"core\_model\_parameter\_type"}\FunctionTok{:} \StringTok{"str"}\FunctionTok{,}
    \DataTypeTok{"lower"}\FunctionTok{:} \DecValTok{0}\FunctionTok{,}
    \DataTypeTok{"upper"}\FunctionTok{:} \DecValTok{2}\FunctionTok{\}}\ErrorTok{,}
\end{Highlighting}
\end{Shaded}

The corresponding entries for the \texttt{Net\_CIFAR10} class are shown
below.

\begin{Shaded}
\begin{Highlighting}[]
\FunctionTok{\{}\DataTypeTok{"Net\_CIFAR10"}\FunctionTok{:}
    \FunctionTok{\{}
        \DataTypeTok{"l1"}\FunctionTok{:} \FunctionTok{\{}
            \DataTypeTok{"type"}\FunctionTok{:} \StringTok{"int"}\FunctionTok{,}
            \DataTypeTok{"default"}\FunctionTok{:} \DecValTok{5}\FunctionTok{,}
            \DataTypeTok{"transform"}\FunctionTok{:} \StringTok{"transform\_power\_2\_int"}\FunctionTok{,}
            \DataTypeTok{"lower"}\FunctionTok{:} \DecValTok{2}\FunctionTok{,}
            \DataTypeTok{"upper"}\FunctionTok{:} \DecValTok{9}\FunctionTok{\},}
        \DataTypeTok{"l2"}\FunctionTok{:} \FunctionTok{\{}
            \DataTypeTok{"type"}\FunctionTok{:} \StringTok{"int"}\FunctionTok{,}
            \DataTypeTok{"default"}\FunctionTok{:} \DecValTok{5}\FunctionTok{,}
            \DataTypeTok{"transform"}\FunctionTok{:} \StringTok{"transform\_power\_2\_int"}\FunctionTok{,}
            \DataTypeTok{"lower"}\FunctionTok{:} \DecValTok{2}\FunctionTok{,}
            \DataTypeTok{"upper"}\FunctionTok{:} \DecValTok{9}\FunctionTok{\},}
        \DataTypeTok{"lr\_mult"}\FunctionTok{:} \FunctionTok{\{}
            \DataTypeTok{"type"}\FunctionTok{:} \StringTok{"float"}\FunctionTok{,}
            \DataTypeTok{"default"}\FunctionTok{:} \FloatTok{1.0}\FunctionTok{,}
            \DataTypeTok{"transform"}\FunctionTok{:} \StringTok{"None"}\FunctionTok{,}
            \DataTypeTok{"lower"}\FunctionTok{:} \FloatTok{0.1}\FunctionTok{,}
            \DataTypeTok{"upper"}\FunctionTok{:} \DecValTok{10}\FunctionTok{\},}
        \DataTypeTok{"batch\_size"}\FunctionTok{:} \FunctionTok{\{}
            \DataTypeTok{"type"}\FunctionTok{:} \StringTok{"int"}\FunctionTok{,}
            \DataTypeTok{"default"}\FunctionTok{:} \DecValTok{4}\FunctionTok{,}
            \DataTypeTok{"transform"}\FunctionTok{:} \StringTok{"transform\_power\_2\_int"}\FunctionTok{,}
            \DataTypeTok{"lower"}\FunctionTok{:} \DecValTok{1}\FunctionTok{,}
            \DataTypeTok{"upper"}\FunctionTok{:} \DecValTok{4}\FunctionTok{\},}
        \DataTypeTok{"epochs"}\FunctionTok{:} \FunctionTok{\{}
            \DataTypeTok{"type"}\FunctionTok{:} \StringTok{"int"}\FunctionTok{,}
            \DataTypeTok{"default"}\FunctionTok{:} \DecValTok{3}\FunctionTok{,}
            \DataTypeTok{"transform"}\FunctionTok{:} \StringTok{"transform\_power\_2\_int"}\FunctionTok{,}
            \DataTypeTok{"lower"}\FunctionTok{:} \DecValTok{1}\FunctionTok{,}
            \DataTypeTok{"upper"}\FunctionTok{:} \DecValTok{4}\FunctionTok{\},}
        \DataTypeTok{"k\_folds"}\FunctionTok{:} \FunctionTok{\{}
            \DataTypeTok{"type"}\FunctionTok{:} \StringTok{"int"}\FunctionTok{,}
            \DataTypeTok{"default"}\FunctionTok{:} \DecValTok{2}\FunctionTok{,}
            \DataTypeTok{"transform"}\FunctionTok{:} \StringTok{"None"}\FunctionTok{,}
            \DataTypeTok{"lower"}\FunctionTok{:} \DecValTok{2}\FunctionTok{,}
            \DataTypeTok{"upper"}\FunctionTok{:} \DecValTok{3}\FunctionTok{\},}
        \DataTypeTok{"patience"}\FunctionTok{:} \FunctionTok{\{}
            \DataTypeTok{"type"}\FunctionTok{:} \StringTok{"int"}\FunctionTok{,}
            \DataTypeTok{"default"}\FunctionTok{:} \DecValTok{5}\FunctionTok{,}
            \DataTypeTok{"transform"}\FunctionTok{:} \StringTok{"None"}\FunctionTok{,}
            \DataTypeTok{"lower"}\FunctionTok{:} \DecValTok{2}\FunctionTok{,}
            \DataTypeTok{"upper"}\FunctionTok{:} \DecValTok{10}\FunctionTok{\},}
        \DataTypeTok{"optimizer"}\FunctionTok{:} \FunctionTok{\{}
            \DataTypeTok{"levels"}\FunctionTok{:} \OtherTok{[}\StringTok{"Adadelta"}\OtherTok{,}
                       \StringTok{"Adagrad"}\OtherTok{,}
                       \StringTok{"Adam"}\OtherTok{,}
                       \StringTok{"AdamW"}\OtherTok{,}
                       \StringTok{"SparseAdam"}\OtherTok{,}
                       \StringTok{"Adamax"}\OtherTok{,}
                       \StringTok{"ASGD"}\OtherTok{,}
                       \StringTok{"LBFGS"}\OtherTok{,}
                       \StringTok{"NAdam"}\OtherTok{,}
                       \StringTok{"RAdam"}\OtherTok{,}
                       \StringTok{"RMSprop"}\OtherTok{,}
                       \StringTok{"Rprop"}\OtherTok{,}
                       \StringTok{"SGD"}\OtherTok{]}\FunctionTok{,}
            \DataTypeTok{"type"}\FunctionTok{:} \StringTok{"factor"}\FunctionTok{,}
            \DataTypeTok{"default"}\FunctionTok{:} \StringTok{"SGD"}\FunctionTok{,}
            \DataTypeTok{"transform"}\FunctionTok{:} \StringTok{"None"}\FunctionTok{,}
            \DataTypeTok{"class\_name"}\FunctionTok{:} \StringTok{"torch.optim"}\FunctionTok{,}
            \DataTypeTok{"core\_model\_parameter\_type"}\FunctionTok{:} \StringTok{"str"}\FunctionTok{,}
            \DataTypeTok{"lower"}\FunctionTok{:} \DecValTok{0}\FunctionTok{,}
            \DataTypeTok{"upper"}\FunctionTok{:} \DecValTok{12}\FunctionTok{\},}
        \DataTypeTok{"sgd\_momentum"}\FunctionTok{:} \FunctionTok{\{}
            \DataTypeTok{"type"}\FunctionTok{:} \StringTok{"float"}\FunctionTok{,}
            \DataTypeTok{"default"}\FunctionTok{:} \FloatTok{0.0}\FunctionTok{,}
            \DataTypeTok{"transform"}\FunctionTok{:} \StringTok{"None"}\FunctionTok{,}
            \DataTypeTok{"lower"}\FunctionTok{:} \FloatTok{0.0}\FunctionTok{,}
            \DataTypeTok{"upper"}\FunctionTok{:} \FloatTok{1.0}\FunctionTok{\}}
    \FunctionTok{\}}
\FunctionTok{\}}
\end{Highlighting}
\end{Shaded}

\hypertarget{sec-modification-of-hyperparameters}{%
\subsection{Modifying the
Hyperparameters}\label{sec-modification-of-hyperparameters}}

Ray tune (PyTorch 2023a) does not provide a way to change the specified
hyperparameters without re-compilation. However, \texttt{spotPython}
provides functions for modifying the hyperparameters, their bounds and
factors as well as for activating and de-activating hyperparameters
without re-compilation of the Python source code. These functions are
described in the following.

\hypertarget{sec-modification-of-default-values}{%
\subsubsection{\texorpdfstring{Modify \texttt{hyper\_dict}
Hyperparameters for the Selected Algorithm aka
\texttt{core\_model}}{Modify hyper\_dict Hyperparameters for the Selected Algorithm aka core\_model}}\label{sec-modification-of-default-values}}

After specifying the model, the corresponding hyperparameters, their
types and bounds are loaded from the \texttt{JSON} file
\texttt{torch\_hyper\_dict.json}. After loading, the user can modify the
hyperparameters, e.g., the bounds. \texttt{spotPython} provides a simple
rule for de-activating hyperparameters: If the lower and the upper bound
are set to identical values, the hyperparameter is de-activated. This is
useful for the hyperparameter tuning, because it allows to specify a
hyperparameter in the \texttt{JSON} file, but to de-activate it in the
\texttt{fun\_control} dictionary. This is done in the next step.

\hypertarget{modify-hyperparameters-of-type-numeric-and-integer-boolean}{%
\subsubsection{Modify Hyperparameters of Type numeric and integer
(boolean)}\label{modify-hyperparameters-of-type-numeric-and-integer-boolean}}

Since the hyperparameter \texttt{k\_folds} is not used in the
\texttt{PyTorch} tutorial, it is de-activated here by setting the lower
and upper bound to the same value. Note, \texttt{k\_folds} is of type
``integer''.

\begin{Shaded}
\begin{Highlighting}[]
\ImportTok{from}\NormalTok{ spotPython.hyperparameters.values }\ImportTok{import}\NormalTok{ modify\_hyper\_parameter\_bounds}
\NormalTok{fun\_control }\OperatorTok{=}\NormalTok{ modify\_hyper\_parameter\_bounds(fun\_control, }
    \StringTok{"batch\_size"}\NormalTok{, bounds}\OperatorTok{=}\NormalTok{[}\DecValTok{1}\NormalTok{, }\DecValTok{5}\NormalTok{])}
\NormalTok{fun\_control }\OperatorTok{=}\NormalTok{ modify\_hyper\_parameter\_bounds(fun\_control, }
    \StringTok{"k\_folds"}\NormalTok{, bounds}\OperatorTok{=}\NormalTok{[}\DecValTok{0}\NormalTok{, }\DecValTok{0}\NormalTok{])}
\NormalTok{fun\_control }\OperatorTok{=}\NormalTok{ modify\_hyper\_parameter\_bounds(fun\_control, }
    \StringTok{"patience"}\NormalTok{, bounds}\OperatorTok{=}\NormalTok{[}\DecValTok{3}\NormalTok{, }\DecValTok{3}\NormalTok{])}
\end{Highlighting}
\end{Shaded}

\hypertarget{modify-hyperparameter-of-type-factor}{%
\subsubsection{Modify Hyperparameter of Type
factor}\label{modify-hyperparameter-of-type-factor}}

In a similar manner as for the numerical hyperparameters, the
categorical hyperparameters can be modified. New configurations can be
chosen by adding or deleting levels. For example, the hyperparameter
\texttt{optimizer} can be re-configured as follows:

In the following setting, two optimizers (\texttt{"SGD"} and
\texttt{"Adam"}) will be compared during the \texttt{spotPython}
hyperparameter tuning. The hyperparameter \texttt{optimizer} is active.

\begin{Shaded}
\begin{Highlighting}[]
\ImportTok{from}\NormalTok{ spotPython.hyperparameters.values }\ImportTok{import}\NormalTok{ modify\_hyper\_parameter\_levels}
\NormalTok{fun\_control }\OperatorTok{=}\NormalTok{ modify\_hyper\_parameter\_levels(fun\_control,}
     \StringTok{"optimizer"}\NormalTok{, [}\StringTok{"SGD"}\NormalTok{, }\StringTok{"Adam"}\NormalTok{])}
\end{Highlighting}
\end{Shaded}

The hyperparameter \texttt{optimizer} can be de-activated by choosing
only one value (level), here: \texttt{"SGD"}.

\begin{Shaded}
\begin{Highlighting}[]
\NormalTok{fun\_control }\OperatorTok{=}\NormalTok{ modify\_hyper\_parameter\_levels(fun\_control, }\StringTok{"optimizer"}\NormalTok{, [}\StringTok{"SGD"}\NormalTok{])}
\end{Highlighting}
\end{Shaded}

As discussed in Section~\ref{sec-optimizers}, there are some issues with
the LBFGS optimizer. Therefore, the usage of the LBFGS optimizer is not
deactivated in \texttt{spotPython} by default. However, the LBFGS
optimizer can be activated by adding it to the list of optimizers.
\texttt{Rprop} was removed, because it does perform very poorly (as some
pre-tests have shown). However, it can also be activated by adding it to
the list of optimizers. Since \texttt{SparseAdam} does not support dense
gradients, \texttt{Adam} was used instead. Therefore, there are 10
default optimizers:

\begin{Shaded}
\begin{Highlighting}[]
\NormalTok{fun\_control }\OperatorTok{=}\NormalTok{ modify\_hyper\_parameter\_levels(fun\_control, }\StringTok{"optimizer"}\NormalTok{,}
\NormalTok{    [}\StringTok{"Adadelta"}\NormalTok{, }\StringTok{"Adagrad"}\NormalTok{, }\StringTok{"Adam"}\NormalTok{, }\StringTok{"AdamW"}\NormalTok{, }\StringTok{"Adamax"}\NormalTok{, }\StringTok{"ASGD"}\NormalTok{, }
    \StringTok{"NAdam"}\NormalTok{, }\StringTok{"RAdam"}\NormalTok{, }\StringTok{"RMSprop"}\NormalTok{, }\StringTok{"SGD"}\NormalTok{])}
\end{Highlighting}
\end{Shaded}

\hypertarget{sec-optimizers}{%
\subsubsection{Optimizers}\label{sec-optimizers}}

Table~\ref{tbl-optimizers} shows some of the optimizers available in
\texttt{PyTorch}:

\hypertarget{tbl-optimizers}{}
\begin{longtable}[]{@{}
  >{\raggedright\arraybackslash}p{(\columnwidth - 28\tabcolsep) * \real{0.1176}}
  >{\raggedright\arraybackslash}p{(\columnwidth - 28\tabcolsep) * \real{0.0588}}
  >{\raggedright\arraybackslash}p{(\columnwidth - 28\tabcolsep) * \real{0.0588}}
  >{\raggedright\arraybackslash}p{(\columnwidth - 28\tabcolsep) * \real{0.0588}}
  >{\raggedright\arraybackslash}p{(\columnwidth - 28\tabcolsep) * \real{0.0588}}
  >{\raggedright\arraybackslash}p{(\columnwidth - 28\tabcolsep) * \real{0.0588}}
  >{\raggedright\arraybackslash}p{(\columnwidth - 28\tabcolsep) * \real{0.0588}}
  >{\raggedright\arraybackslash}p{(\columnwidth - 28\tabcolsep) * \real{0.0588}}
  >{\raggedright\arraybackslash}p{(\columnwidth - 28\tabcolsep) * \real{0.0588}}
  >{\raggedright\arraybackslash}p{(\columnwidth - 28\tabcolsep) * \real{0.0735}}
  >{\raggedright\arraybackslash}p{(\columnwidth - 28\tabcolsep) * \real{0.0735}}
  >{\raggedright\arraybackslash}p{(\columnwidth - 28\tabcolsep) * \real{0.0735}}
  >{\raggedright\arraybackslash}p{(\columnwidth - 28\tabcolsep) * \real{0.0735}}
  >{\raggedright\arraybackslash}p{(\columnwidth - 28\tabcolsep) * \real{0.0588}}
  >{\raggedright\arraybackslash}p{(\columnwidth - 28\tabcolsep) * \real{0.0588}}@{}}
\caption{\label{tbl-optimizers}Optimizers available in PyTorch
(selection). ``mom'' denotes \texttt{momentum}, ``weight''
\texttt{weight\_decay}, ``damp'' \texttt{dampening}, ``nest''
\texttt{nesterov}, ``lr\_sc''
\texttt{learning\ rate\ for\ scaling\ delta}, ``mom\_dec'' for
\texttt{momentum\_decay}, and ``step\_s'' for \texttt{step\_sizes}. The
default values are shown in the table.}\tabularnewline
\toprule\noalign{}
\begin{minipage}[b]{\linewidth}\raggedright
Optimizer
\end{minipage} & \begin{minipage}[b]{\linewidth}\raggedright
lr
\end{minipage} & \begin{minipage}[b]{\linewidth}\raggedright
mom
\end{minipage} & \begin{minipage}[b]{\linewidth}\raggedright
weight
\end{minipage} & \begin{minipage}[b]{\linewidth}\raggedright
damp
\end{minipage} & \begin{minipage}[b]{\linewidth}\raggedright
nest
\end{minipage} & \begin{minipage}[b]{\linewidth}\raggedright
rho
\end{minipage} & \begin{minipage}[b]{\linewidth}\raggedright
lr\_sc
\end{minipage} & \begin{minipage}[b]{\linewidth}\raggedright
lr\_decay
\end{minipage} & \begin{minipage}[b]{\linewidth}\raggedright
betas
\end{minipage} & \begin{minipage}[b]{\linewidth}\raggedright
lambd
\end{minipage} & \begin{minipage}[b]{\linewidth}\raggedright
alpha
\end{minipage} & \begin{minipage}[b]{\linewidth}\raggedright
mom\_decay
\end{minipage} & \begin{minipage}[b]{\linewidth}\raggedright
etas
\end{minipage} & \begin{minipage}[b]{\linewidth}\raggedright
step\_s
\end{minipage} \\
\midrule\noalign{}
\endfirsthead
\toprule\noalign{}
\begin{minipage}[b]{\linewidth}\raggedright
Optimizer
\end{minipage} & \begin{minipage}[b]{\linewidth}\raggedright
lr
\end{minipage} & \begin{minipage}[b]{\linewidth}\raggedright
mom
\end{minipage} & \begin{minipage}[b]{\linewidth}\raggedright
weight
\end{minipage} & \begin{minipage}[b]{\linewidth}\raggedright
damp
\end{minipage} & \begin{minipage}[b]{\linewidth}\raggedright
nest
\end{minipage} & \begin{minipage}[b]{\linewidth}\raggedright
rho
\end{minipage} & \begin{minipage}[b]{\linewidth}\raggedright
lr\_sc
\end{minipage} & \begin{minipage}[b]{\linewidth}\raggedright
lr\_decay
\end{minipage} & \begin{minipage}[b]{\linewidth}\raggedright
betas
\end{minipage} & \begin{minipage}[b]{\linewidth}\raggedright
lambd
\end{minipage} & \begin{minipage}[b]{\linewidth}\raggedright
alpha
\end{minipage} & \begin{minipage}[b]{\linewidth}\raggedright
mom\_decay
\end{minipage} & \begin{minipage}[b]{\linewidth}\raggedright
etas
\end{minipage} & \begin{minipage}[b]{\linewidth}\raggedright
step\_s
\end{minipage} \\
\midrule\noalign{}
\endhead
\bottomrule\noalign{}
\endlastfoot
Adadelta & - & - & 0. & - & - & 0.9 & 1.0 & - & - & - & - & - & - & - \\
Adagrad & 1e-2 & - & 0. & - & - & - & - & 0. & - & - & - & - & - & - \\
Adam & 1e-3 & - & 0. & - & - & - & - & - & (0.9,0.999) & - & - & - & - &
- \\
AdamW & 1e-3 & - & 1e-2 & - & - & - & - & - & (0.9,0.999) & - & - & - &
- & - \\
SparseAdam & 1e-3 & - & - & - & - & - & - & - & (0.9,0.999) & - & - & -
& - & - \\
Adamax & 2e-3 & - & 0. & - & - & - & - & - & (0.9, 0.999) & - & - & - &
- & - \\
ASGD & 1e-2 & 0.9 & 0. & - & False & - & - & - & - & 1e-4 & 0.75 & - & -
& - \\
LBFGS & 1. & - & - & - & - & - & - & - & - & - & - & - & - & - \\
NAdam & 2e-3 & - & 0. & - & - & - & - & - & (0.9,0.999) & - & - & 0 & -
& - \\
RAdam & 1e-3 & - & 0. & - & - & - & - & - & (0.9,0.999) & - & - & - & -
& - \\
RMSprop & 1e-2 & 0. & 0. & - & - & - & - & - & (0.9,0.999) & - & - & - &
- & - \\
Rprop & 1e-2 & - & - & - & - & - & - & - & - & - & (0.5,1.2) & (1e-6,
50) & - & - \\
SGD & required & 0. & 0. & 0. & False & - & - & - & - & - & - & - & - &
- \\
\end{longtable}

\texttt{spotPython} implements an \texttt{optimization} handler that
maps the optimizer names to the corresponding \texttt{PyTorch}
optimizers.

\begin{tcolorbox}[enhanced jigsaw, colbacktitle=quarto-callout-note-color!10!white, coltitle=black, opacitybacktitle=0.6, title=\textcolor{quarto-callout-note-color}{\faInfo}\hspace{0.5em}{A note on LBFGS}, bottomrule=.15mm, arc=.35mm, leftrule=.75mm, breakable, colframe=quarto-callout-note-color-frame, colback=white, left=2mm, toptitle=1mm, bottomtitle=1mm, rightrule=.15mm, titlerule=0mm, toprule=.15mm, opacityback=0]

We recommend deactivating \texttt{PyTorch}'s LBFGS optimizer, because it
does not perform very well. The \texttt{PyTorch} documentation, see
\url{https://pytorch.org/docs/stable/generated/torch.optim.LBFGS.html\#torch.optim.LBFGS},
states:

\begin{quote}
This is a very memory intensive optimizer (it requires additional
\texttt{param\_bytes\ *\ (history\_size\ +\ 1)} bytes). If it doesn't
fit in memory try reducing the history size, or use a different
algorithm.
\end{quote}

Furthermore, the LBFGS optimizer is not compatible with the
\texttt{PyTorch} tutorial. The reason is that the LBFGS optimizer
requires the \texttt{closure} function, which is not implemented in the
\texttt{PyTorch} tutorial. Therefore, the \texttt{LBFGS} optimizer is
recommended here.

\end{tcolorbox}

Since there are 10 optimizers in the portfolio, it is not recommended
tuning the hyperparameters that effect one single optimizer only.

\begin{tcolorbox}[enhanced jigsaw, colbacktitle=quarto-callout-note-color!10!white, coltitle=black, opacitybacktitle=0.6, title=\textcolor{quarto-callout-note-color}{\faInfo}\hspace{0.5em}{A note on the learning rate}, bottomrule=.15mm, arc=.35mm, leftrule=.75mm, breakable, colframe=quarto-callout-note-color-frame, colback=white, left=2mm, toptitle=1mm, bottomtitle=1mm, rightrule=.15mm, titlerule=0mm, toprule=.15mm, opacityback=0]

\texttt{spotPython} provides a multiplier for the default learning
rates, \texttt{lr\_mult}, because optimizers use different learning
rates. Using a multiplier for the learning rates might enable a
simultaneous tuning of the learning rates for all optimizers. However,
this is not recommended, because the learning rates are not comparable
across optimizers. Therefore, we recommend fixing the learning rate for
all optimizers if multiple optimizers are used. This can be done by
setting the lower and upper bounds of the learning rate multiplier to
the same value as shown below.

\end{tcolorbox}

Thus, the learning rate, which affects the \texttt{SGD} optimizer, will
be set to a fixed value. We choose the default value of \texttt{1e-3}
for the learning rate, because it is used in other \texttt{PyTorch}
examples (it is also the default value used by \texttt{spotPython} as
defined in the \texttt{optimizer\_handler()} method). We recommend
tuning the learning rate later, when a reduced set of optimizers is
fixed. Here, we will demonstrate how to select in a screening phase the
optimizers that should be used for the hyperparameter tuning.

For the same reason, we will fix the \texttt{sgd\_momentum} to
\texttt{0.9}.

\begin{Shaded}
\begin{Highlighting}[]
\NormalTok{fun\_control }\OperatorTok{=}\NormalTok{ modify\_hyper\_parameter\_bounds(fun\_control, }\StringTok{"lr\_mult"}\NormalTok{,}
\NormalTok{    bounds}\OperatorTok{=}\NormalTok{[}\FloatTok{1.0}\NormalTok{, }\FloatTok{1.0}\NormalTok{])}
\NormalTok{fun\_control }\OperatorTok{=}\NormalTok{ modify\_hyper\_parameter\_bounds(fun\_control, }\StringTok{"sgd\_momentum"}\NormalTok{,}
\NormalTok{    bounds}\OperatorTok{=}\NormalTok{[}\FloatTok{0.9}\NormalTok{, }\FloatTok{0.9}\NormalTok{])}
\end{Highlighting}
\end{Shaded}

\hypertarget{sec-selection-of-target-function}{%
\subsection{Evaluation}\label{sec-selection-of-target-function}}

The evaluation procedure requires the specification of two elements:

\begin{enumerate}
\def\labelenumi{\arabic{enumi}.}
\tightlist
\item
  the way how the data is split into a train and a test set and
\item
  the loss function (and a metric).
\end{enumerate}

\hypertarget{hold-out-data-split-and-cross-validation}{%
\subsubsection{Hold-out Data Split and
Cross-Validation}\label{hold-out-data-split-and-cross-validation}}

As a default, \texttt{spotPython} provides a standard hold-out data
split and cross validation.

\hypertarget{hold-out-data-split}{%
\paragraph{Hold-out Data Split}\label{hold-out-data-split}}

If a hold-out data split is used, the data will be partitioned into a
training, a validation, and a test data set. The split depends on the
setting of the \texttt{eval} parameter. If \texttt{eval} is set to
\texttt{train\_hold\_out}, one data set, usually the original training
data set, is split into a new training and a validation data set. The
training data set is used for training the model. The validation data
set is used for the evaluation of the hyperparameter configuration and
early stopping to prevent overfitting. In this case, the original test
data set is not used. The following splits are performed in the hold-out
setting:
\(\{\text{train}_0, \text{test}\} \rightarrow \{\text{train}_1, \text{validation}_1, \text{test}\}\),
where \(\text{train}_1 \cup \text{validation}_1 = \text{train}_0\).

\begin{tcolorbox}[enhanced jigsaw, colbacktitle=quarto-callout-note-color!10!white, coltitle=black, opacitybacktitle=0.6, title=\textcolor{quarto-callout-note-color}{\faInfo}\hspace{0.5em}{Note}, bottomrule=.15mm, arc=.35mm, leftrule=.75mm, breakable, colframe=quarto-callout-note-color-frame, colback=white, left=2mm, toptitle=1mm, bottomtitle=1mm, rightrule=.15mm, titlerule=0mm, toprule=.15mm, opacityback=0]

\texttt{spotPython} returns the hyperparameters of the machine learning
and deep learning models, e.g., number of layers, learning rate, or
optimizer, but not the model weights. Therefore, after the SPOT run is
finished, the corresponding model with the optimized architecture has to
be trained again with the best hyperparameter configuration. The
training is performed on the training data set. The test data set is
used for the final evaluation of the model.

Summarizing, the following splits are performed in the hold-out setting:

\begin{enumerate}
\def\labelenumi{\arabic{enumi}.}
\tightlist
\item
  Run \texttt{spotPython} with \texttt{eval} set to
  \texttt{train\_hold\_out} to determine the best hyperparameter
  configuration.
\item
  Train the model with the best hyperparameter configuration
  (``architecture'') on the training data set:

  \begin{itemize}
  \tightlist
  \item
    \texttt{train\_tuned(model\_spot,\ train,\ "model\_spot.pt")}.
  \end{itemize}
\item
  Test the model on the test data:

  \begin{itemize}
  \tightlist
  \item
    \texttt{test\_tuned(model\_spot,\ test,\ "model\_spot.pt")}
  \end{itemize}
\end{enumerate}

These steps will be exemplified in the following sections.

\end{tcolorbox}

In addition to this \texttt{hold-out} setting, \texttt{spotPython}
provides another hold-out setting, where an explicit test data is
specified by the user that will be used as the validation set. To choose
this option, the \texttt{eval} parameter is set to
\texttt{test\_hold\_out}. In this case, the training data set is used
for the model training. Then, the explicitly defined test data set is
used for the evaluation of the hyperparameter configuration (the
validation).

\hypertarget{cross-validation}{%
\paragraph{Cross-Validation}\label{cross-validation}}

The cross validation setting is used by setting the \texttt{eval}
parameter to \texttt{train\_cv} or \texttt{test\_cv}. In both cases, the
data set is split into \(k\) folds. The model is trained on \(k-1\)
folds and evaluated on the remaining fold. This is repeated \(k\) times,
so that each fold is used exactly once for evaluation. The final
evaluation is performed on the test data set. The cross validation
setting is useful for small data sets, because it allows to use all data
for training and evaluation. However, it is computationally expensive,
because the model has to be trained \(k\) times.

\begin{tcolorbox}[enhanced jigsaw, colbacktitle=quarto-callout-note-color!10!white, coltitle=black, opacitybacktitle=0.6, title=\textcolor{quarto-callout-note-color}{\faInfo}\hspace{0.5em}{Note}, bottomrule=.15mm, arc=.35mm, leftrule=.75mm, breakable, colframe=quarto-callout-note-color-frame, colback=white, left=2mm, toptitle=1mm, bottomtitle=1mm, rightrule=.15mm, titlerule=0mm, toprule=.15mm, opacityback=0]

Combinations of the above settings are possible, e.g., cross validation
can be used for training and hold-out for evaluation or \emph{vice
versa}. Also, cross validation can be used for training and testing.
Because cross validation is not used in the \texttt{PyTorch} tutorial
(PyTorch 2023a), it is not considered further here.

\end{tcolorbox}

\hypertarget{overview-of-the-evaluation-settings}{%
\paragraph{Overview of the Evaluation
Settings}\label{overview-of-the-evaluation-settings}}

\hypertarget{settings-for-the-hyperparameter-tuning}{%
\subparagraph{Settings for the Hyperparameter
Tuning}\label{settings-for-the-hyperparameter-tuning}}

Table~\ref{tbl-eval-settings} provides an overview of the training
evaluations.

\hypertarget{tbl-eval-settings}{}
\begin{longtable}[]{@{}
  >{\raggedright\arraybackslash}p{(\columnwidth - 8\tabcolsep) * \real{0.1429}}
  >{\centering\arraybackslash}p{(\columnwidth - 8\tabcolsep) * \real{0.1429}}
  >{\centering\arraybackslash}p{(\columnwidth - 8\tabcolsep) * \real{0.1429}}
  >{\raggedright\arraybackslash}p{(\columnwidth - 8\tabcolsep) * \real{0.2857}}
  >{\raggedright\arraybackslash}p{(\columnwidth - 8\tabcolsep) * \real{0.2857}}@{}}
\caption{\label{tbl-eval-settings}Overview of the evaluation
settings.}\tabularnewline
\toprule\noalign{}
\begin{minipage}[b]{\linewidth}\raggedright
\texttt{eval}
\end{minipage} & \begin{minipage}[b]{\linewidth}\centering
\texttt{train}
\end{minipage} & \begin{minipage}[b]{\linewidth}\centering
\texttt{test}
\end{minipage} & \begin{minipage}[b]{\linewidth}\raggedright
function
\end{minipage} & \begin{minipage}[b]{\linewidth}\raggedright
comment
\end{minipage} \\
\midrule\noalign{}
\endfirsthead
\toprule\noalign{}
\begin{minipage}[b]{\linewidth}\raggedright
\texttt{eval}
\end{minipage} & \begin{minipage}[b]{\linewidth}\centering
\texttt{train}
\end{minipage} & \begin{minipage}[b]{\linewidth}\centering
\texttt{test}
\end{minipage} & \begin{minipage}[b]{\linewidth}\raggedright
function
\end{minipage} & \begin{minipage}[b]{\linewidth}\raggedright
comment
\end{minipage} \\
\midrule\noalign{}
\endhead
\bottomrule\noalign{}
\endlastfoot
\texttt{"train\_hold\_out"} & \(\checkmark\) & &
\texttt{train\_one\_epoch()}, \texttt{validate\_one\_epoch()} for early
stopping & splits the \texttt{train} data set internally \\
\texttt{"test\_hold\_out"} & \(\checkmark\) & \(\checkmark\) &
\texttt{train\_one\_epoch()}, \texttt{validate\_one\_epoch()} for early
stopping & use the \texttt{test\ data\ set} for
\texttt{validate\_one\_epoch()} \\
\texttt{"train\_cv"} & \(\checkmark\) & &
\texttt{evaluate\_cv(net,\ train)} & CV using the \texttt{train} data
set \\
\texttt{"test\_cv"} & & \(\checkmark\) &
\texttt{evaluate\_cv(net,\ test)} & CV using the \texttt{test} data set
. Identical to \texttt{"train\_cv"}, uses only test data. \\
\end{longtable}

\begin{itemize}
\tightlist
\item
  \texttt{"train\_cv"} and \texttt{"test\_cv"} use
  \texttt{sklearn.model\_selection.KFold()} internally.
\end{itemize}

Section~\ref{sec-data-splitting} (in the Appendix) provides more details
on the data splitting.

\hypertarget{settings-for-the-final-evaluation-of-the-tuned-architecture}{%
\paragraph{Settings for the Final Evaluation of the Tuned
Architecture}\label{settings-for-the-final-evaluation-of-the-tuned-architecture}}

\hypertarget{training-of-the-tuned-architecture}{%
\subparagraph{Training of the Tuned
Architecture}\label{training-of-the-tuned-architecture}}

\texttt{train\_tuned(model,\ train)}: train the model with the best
hyperparameter configuration (or simply the default) on the training
data set. It splits the \texttt{train}data into new \texttt{train} and
\texttt{validation} sets using
\texttt{create\_train\_val\_data\_loaders()}, which calls
\texttt{torch.utils.data.random\_split()} internally. Currently, 60\% of
the data is used for training and 40\% for validation. The
\texttt{train} data is used for training the model with
\texttt{train\_hold\_out()}. The \texttt{validation} data is used for
early stopping using \texttt{validate\_fold\_or\_hold\_out()} on the
\texttt{validation} data set.

\hypertarget{testing-of-the-tuned-architecture}{%
\subparagraph{Testing of the Tuned
Architecture}\label{testing-of-the-tuned-architecture}}

\texttt{test\_tuned(model,\ test)}: test the model on the test data set.
No data splitting is performed. The (trained) model is evaluated using
the \texttt{validate\_fold\_or\_hold\_out()} function.

Note: During training, \texttt{shuffle} is set to \texttt{True}, whereas
during testing, \texttt{shuffle} is set to \texttt{False}.

Section~\ref{sec-final-model-evaluation} describes the final evaluation
of the tuned architecture.

\hypertarget{loss-functions-and-metrics}{%
\subsubsection{Loss Functions and
Metrics}\label{loss-functions-and-metrics}}

The key \texttt{"loss\_function"} specifies the loss function which is
used during the optimization. There are several different loss functions
under \texttt{PyTorch}'s \texttt{nn} package. For example, a simple loss
is \texttt{MSELoss}, which computes the mean-squared error between the
output and the target. In this tutorial we will use
\texttt{CrossEntropyLoss}, because it is also used in the
\texttt{PyTorch} tutorial.

\begin{Shaded}
\begin{Highlighting}[]
\ImportTok{from}\NormalTok{ torch.nn }\ImportTok{import}\NormalTok{ CrossEntropyLoss}
\NormalTok{loss\_function }\OperatorTok{=}\NormalTok{ CrossEntropyLoss()}
\NormalTok{fun\_control.update(\{}\StringTok{"loss\_function"}\NormalTok{: loss\_function\})}
\end{Highlighting}
\end{Shaded}

In addition to the loss functions, \texttt{spotPython} provides access
to a large number of metrics.

\begin{itemize}
\tightlist
\item
  The key \texttt{"metric\_sklearn"} is used for metrics that follow the
  \texttt{scikit-learn} conventions.
\item
  The key \texttt{"river\_metric"} is used for the river based
  evaluation (Montiel et al. 2021) via
  \texttt{eval\_oml\_iter\_progressive}, and
\item
  the key \texttt{"metric\_torch"} is used for the metrics from
  \texttt{TorchMetrics}.
\end{itemize}

\texttt{TorchMetrics} is a collection of more than 90 PyTorch
metrics\footnote{\href{https://torchmetrics.readthedocs.io/en/latest/}{https://torchmetrics.readthedocs.io/en/latest/.}}.

Because the \texttt{PyTorch} tutorial uses the accuracy as metric, we
use the same metric here. Currently, accuracy is computed in the
tutorial's example code. We will use \texttt{TorchMetrics} instead,
because it offers more flexibilty, e.g., it can be used for regression
and classification. Furthermore, \texttt{TorchMetrics} offers the
following advantages:

\begin{itemize}
\tightlist
\item
  A standardized interface to increase reproducibility
\item
  Reduces Boilerplate
\item
  Distributed-training compatible
\item
  Rigorously tested
\item
  Automatic accumulation over batches
\item
  Automatic synchronization between multiple devices
\end{itemize}

Therefore, we set

\begin{Shaded}
\begin{Highlighting}[]
\ImportTok{import}\NormalTok{ torchmetrics}
\NormalTok{metric\_torch }\OperatorTok{=}\NormalTok{ torchmetrics.Accuracy(task}\OperatorTok{=}\StringTok{"multiclass"}\NormalTok{, num\_classes}\OperatorTok{=}\DecValTok{10}\NormalTok{)}
\end{Highlighting}
\end{Shaded}

\begin{Shaded}
\begin{Highlighting}[]
\NormalTok{loss\_function }\OperatorTok{=}\NormalTok{ CrossEntropyLoss()}
\NormalTok{weights }\OperatorTok{=} \FloatTok{1.0}
\NormalTok{metric\_torch }\OperatorTok{=}\NormalTok{ torchmetrics.Accuracy(task}\OperatorTok{=}\StringTok{"multiclass"}\NormalTok{, num\_classes}\OperatorTok{=}\DecValTok{10}\NormalTok{)}
\NormalTok{shuffle }\OperatorTok{=} \VariableTok{True}
\BuiltInTok{eval} \OperatorTok{=} \StringTok{"train\_hold\_out"}
\NormalTok{device }\OperatorTok{=}\NormalTok{ DEVICE}
\NormalTok{show\_batch\_interval }\OperatorTok{=} \DecValTok{100\_000}
\NormalTok{path}\OperatorTok{=}\StringTok{"torch\_model.pt"}

\NormalTok{fun\_control.update(\{}
               \StringTok{"data\_dir"}\NormalTok{: }\VariableTok{None}\NormalTok{,}
               \StringTok{"checkpoint\_dir"}\NormalTok{: }\VariableTok{None}\NormalTok{,}
               \StringTok{"horizon"}\NormalTok{: }\VariableTok{None}\NormalTok{,}
               \StringTok{"oml\_grace\_period"}\NormalTok{: }\VariableTok{None}\NormalTok{,}
               \StringTok{"weights"}\NormalTok{: weights,}
               \StringTok{"step"}\NormalTok{: }\VariableTok{None}\NormalTok{,}
               \StringTok{"log\_level"}\NormalTok{: }\DecValTok{50}\NormalTok{,}
               \StringTok{"weight\_coeff"}\NormalTok{: }\VariableTok{None}\NormalTok{,}
               \StringTok{"metric\_torch"}\NormalTok{: metric\_torch,}
               \StringTok{"metric\_river"}\NormalTok{: }\VariableTok{None}\NormalTok{,}
               \StringTok{"metric\_sklearn"}\NormalTok{: }\VariableTok{None}\NormalTok{,}
               \StringTok{"loss\_function"}\NormalTok{: loss\_function,}
               \StringTok{"shuffle"}\NormalTok{: shuffle,}
               \StringTok{"eval"}\NormalTok{: }\BuiltInTok{eval}\NormalTok{,}
               \StringTok{"device"}\NormalTok{: device,}
               \StringTok{"show\_batch\_interval"}\NormalTok{: show\_batch\_interval,}
               \StringTok{"path"}\NormalTok{: path,}
\NormalTok{               \})}
\end{Highlighting}
\end{Shaded}

\hypertarget{sec-call-the-hyperparameter-tuner}{%
\subsection{Calling the SPOT
Function}\label{sec-call-the-hyperparameter-tuner}}

Now, the dictionary \texttt{fun\_control} contains all information
needed for the hyperparameter tuning. Before the hyperparameter tuning
is started, it is recommended to take a look at the experimental design.
The method \texttt{gen\_design\_table} generates a design table as
follows:

\begin{Shaded}
\begin{Highlighting}[]
\ImportTok{from}\NormalTok{ spotPython.utils.eda }\ImportTok{import}\NormalTok{ gen\_design\_table}
\BuiltInTok{print}\NormalTok{(gen\_design\_table(fun\_control))}
\end{Highlighting}
\end{Shaded}

This allows to check if all information is available and if the
information is correct. Table~\ref{tbl-design} shows the experimental
design for the hyperparameter tuning. Hyperparameter transformations are
shown in the column ``transform'', e.g., the \texttt{l1} default is
\texttt{5}, which results in the value \(2^5 = 32\) for the network,
because the transformation \texttt{transform\_power\_2\_int} was
selected in the \texttt{JSON} file. The default value of the
\texttt{batch\_size} is set to \texttt{4}, which results in a batch size
of \(2^4 = 16\).

\hypertarget{tbl-design}{}
\begin{longtable}[]{@{}
  >{\raggedright\arraybackslash}p{(\columnwidth - 10\tabcolsep) * \real{0.1519}}
  >{\raggedright\arraybackslash}p{(\columnwidth - 10\tabcolsep) * \real{0.1013}}
  >{\raggedright\arraybackslash}p{(\columnwidth - 10\tabcolsep) * \real{0.2278}}
  >{\raggedright\arraybackslash}p{(\columnwidth - 10\tabcolsep) * \real{0.1139}}
  >{\raggedright\arraybackslash}p{(\columnwidth - 10\tabcolsep) * \real{0.1139}}
  >{\raggedright\arraybackslash}p{(\columnwidth - 10\tabcolsep) * \real{0.2911}}@{}}
\caption{\label{tbl-design}Experimental design for the hyperparameter
tuning. The table shows the hyperparameters, their types, default
values, lower and upper bounds, and the transformation function. The
transformation function is used to transform the hyperparameter values
from the unit hypercube to the original domain. The transformation
function is applied to the hyperparameter values before the evaluation
of the objective function.}\tabularnewline
\toprule\noalign{}
\begin{minipage}[b]{\linewidth}\raggedright
name
\end{minipage} & \begin{minipage}[b]{\linewidth}\raggedright
type
\end{minipage} & \begin{minipage}[b]{\linewidth}\raggedright
default
\end{minipage} & \begin{minipage}[b]{\linewidth}\raggedright
lower
\end{minipage} & \begin{minipage}[b]{\linewidth}\raggedright
upper
\end{minipage} & \begin{minipage}[b]{\linewidth}\raggedright
transform
\end{minipage} \\
\midrule\noalign{}
\endfirsthead
\toprule\noalign{}
\begin{minipage}[b]{\linewidth}\raggedright
name
\end{minipage} & \begin{minipage}[b]{\linewidth}\raggedright
type
\end{minipage} & \begin{minipage}[b]{\linewidth}\raggedright
default
\end{minipage} & \begin{minipage}[b]{\linewidth}\raggedright
lower
\end{minipage} & \begin{minipage}[b]{\linewidth}\raggedright
upper
\end{minipage} & \begin{minipage}[b]{\linewidth}\raggedright
transform
\end{minipage} \\
\midrule\noalign{}
\endhead
\bottomrule\noalign{}
\endlastfoot
l1 & int & 5 & 2 & 9 & transform\_power\_2\_int \\
l2 & int & 5 & 2 & 9 & transform\_power\_2\_int \\
lr\_mult & float & 1.0 & 1 & 1 & None \\
batch\_size & int & 4 & 1 & 5 & transform\_power\_2\_int \\
epochs & int & 3 & 3 & 4 & transform\_power\_2\_int \\
k\_folds & int & 1 & 0 & 0 & None \\
patience & int & 5 & 3 & 3 & None \\
optimizer & factor & SGD & 0 & 9 & None \\
sgd\_momentum & float & 0.0 & 0.9 & 0.9 & None \\
\end{longtable}

The objective function \texttt{fun\_torch} is selected next. It
implements an interface from \texttt{PyTorch}'s training, validation,
and testing methods to \texttt{spotPython}.

\begin{Shaded}
\begin{Highlighting}[]
\ImportTok{from}\NormalTok{ spotPython.fun.hypertorch }\ImportTok{import}\NormalTok{ HyperTorch}
\NormalTok{fun }\OperatorTok{=}\NormalTok{ HyperTorch().fun\_torch}
\end{Highlighting}
\end{Shaded}

The \texttt{spotPython} hyperparameter tuning is started by calling the
\texttt{Spot} function. Here, we will run the tuner for approximately 30
minutes (\texttt{max\_time}). Note: the initial design is always
evaluated in the \texttt{spotPython} run. As a consequence, the run may
take longer than specified by \texttt{max\_time}, because the evaluation
time of initial design (here: \texttt{init\_size}, 10 points) is
performed independently of \texttt{max\_time}.

\begin{Shaded}
\begin{Highlighting}[]
\ImportTok{from}\NormalTok{ spotPython.spot }\ImportTok{import}\NormalTok{ spot}
\ImportTok{from}\NormalTok{ math }\ImportTok{import}\NormalTok{ inf}
\ImportTok{import}\NormalTok{ numpy }\ImportTok{as}\NormalTok{ np}
\NormalTok{spot\_tuner }\OperatorTok{=}\NormalTok{ spot.Spot(fun}\OperatorTok{=}\NormalTok{fun,}
\NormalTok{                   lower }\OperatorTok{=}\NormalTok{ lower,}
\NormalTok{                   upper }\OperatorTok{=}\NormalTok{ upper,}
\NormalTok{                   fun\_evals }\OperatorTok{=}\NormalTok{ inf,}
\NormalTok{                   fun\_repeats }\OperatorTok{=} \DecValTok{1}\NormalTok{,}
\NormalTok{                   max\_time }\OperatorTok{=}\NormalTok{ MAX\_TIME,}
\NormalTok{                   noise }\OperatorTok{=} \VariableTok{False}\NormalTok{,}
\NormalTok{                   tolerance\_x }\OperatorTok{=}\NormalTok{ np.sqrt(np.spacing(}\DecValTok{1}\NormalTok{)),}
\NormalTok{                   var\_type }\OperatorTok{=}\NormalTok{ var\_type,}
\NormalTok{                   var\_name }\OperatorTok{=}\NormalTok{ var\_name,}
\NormalTok{                   infill\_criterion }\OperatorTok{=} \StringTok{"y"}\NormalTok{,}
\NormalTok{                   n\_points }\OperatorTok{=} \DecValTok{1}\NormalTok{,}
\NormalTok{                   seed}\OperatorTok{=}\DecValTok{123}\NormalTok{,}
\NormalTok{                   log\_level }\OperatorTok{=} \DecValTok{50}\NormalTok{,}
\NormalTok{                   show\_models}\OperatorTok{=} \VariableTok{False}\NormalTok{,}
\NormalTok{                   show\_progress}\OperatorTok{=} \VariableTok{True}\NormalTok{,}
\NormalTok{                   fun\_control }\OperatorTok{=}\NormalTok{ fun\_control,}
\NormalTok{                   design\_control}\OperatorTok{=}\NormalTok{\{}\StringTok{"init\_size"}\NormalTok{: INIT\_SIZE,}
                                   \StringTok{"repeats"}\NormalTok{: }\DecValTok{1}\NormalTok{\},}
\NormalTok{                   surrogate\_control}\OperatorTok{=}\NormalTok{\{}\StringTok{"noise"}\NormalTok{: }\VariableTok{True}\NormalTok{,}
                                      \StringTok{"cod\_type"}\NormalTok{: }\StringTok{"norm"}\NormalTok{,}
                                      \StringTok{"min\_theta"}\NormalTok{: }\OperatorTok{{-}}\DecValTok{4}\NormalTok{,}
                                      \StringTok{"max\_theta"}\NormalTok{: }\DecValTok{3}\NormalTok{,}
                                      \StringTok{"n\_theta"}\NormalTok{: }\BuiltInTok{len}\NormalTok{(var\_name),}
                                      \StringTok{"model\_fun\_evals"}\NormalTok{: }\DecValTok{10\_000}\NormalTok{,}
                                      \StringTok{"log\_level"}\NormalTok{: }\DecValTok{50}
\NormalTok{                                      \})}
\NormalTok{spot\_tuner.run(X\_start}\OperatorTok{=}\NormalTok{X\_start)}
\end{Highlighting}
\end{Shaded}

During the run, the following output is shown:

\begin{Shaded}
\begin{Highlighting}[]
\NormalTok{config: \{\textquotesingle{}l1\textquotesingle{}: 128, \textquotesingle{}l2\textquotesingle{}: 8, \textquotesingle{}lr\_mult\textquotesingle{}: 1.0, \textquotesingle{}batch\_size\textquotesingle{}: 32,}
\NormalTok{    \textquotesingle{}epochs\textquotesingle{}: 16, \textquotesingle{}k\_folds\textquotesingle{}: 0, \textquotesingle{}patience\textquotesingle{}: 3,}
\NormalTok{    \textquotesingle{}optimizer\textquotesingle{}: \textquotesingle{}AdamW\textquotesingle{}, \textquotesingle{}sgd\_momentum\textquotesingle{}: 0.9\}}
\NormalTok{Epoch: 1}
\NormalTok{Loss on hold{-}out set: 1.5143253986358642}
\NormalTok{Accuracy on hold{-}out set: 0.4447}
\NormalTok{MulticlassAccuracy value on hold{-}out data: 0.4447000026702881}
\NormalTok{Epoch: 2}
\NormalTok{...}
\NormalTok{Epoch: 15}
\NormalTok{Loss on hold{-}out set: 1.2061678514480592}
\NormalTok{Accuracy on hold{-}out set: 0.59505}
\NormalTok{MulticlassAccuracy value on hold{-}out data: 0.5950499773025513}
\NormalTok{Early stopping at epoch 14}
\NormalTok{Returned to Spot: Validation loss: 1.2061678514480592}
\NormalTok{{-}{-}{-}{-}{-}{-}{-}{-}{-}{-}{-}{-}{-}{-}{-}{-}{-}{-}{-}{-}{-}{-}{-}{-}{-}{-}{-}{-}{-}{-}{-}{-}{-}{-}{-}{-}{-}{-}{-}{-}{-}{-}{-}{-}{-}{-}}
\end{Highlighting}
\end{Shaded}

\hypertarget{sec-tensorboard}{%
\subsection{Tensorboard}\label{sec-tensorboard}}

The textual output shown in the console (or code cell) can be visualized
with Tensorboard.

\hypertarget{tensorboard-start-tensorboard}{%
\subsubsection{Tensorboard: Start
Tensorboard}\label{tensorboard-start-tensorboard}}

Start TensorBoard through the command line to visualize data you logged.
Specify the root log directory as used in
\texttt{fun\_control\ =\ fun\_control\_init(task="regression",\ tensorboard\_path="runs/24\_spot\_torch\_regression")}
as the \texttt{tensorboard\_path}. The argument logdir points to
directory where TensorBoard will look to find event files that it can
display. TensorBoard will recursively walk the directory structure
rooted at logdir, looking for .\emph{tfevents.} files.

\begin{Shaded}
\begin{Highlighting}[]
\NormalTok{tensorboard {-}{-}logdir=runs}
\end{Highlighting}
\end{Shaded}

Go to the URL it provides or to \url{http://localhost:6006/}. The
following figures show some screenshots of Tensorboard.

\begin{figure}

{\centering \includegraphics{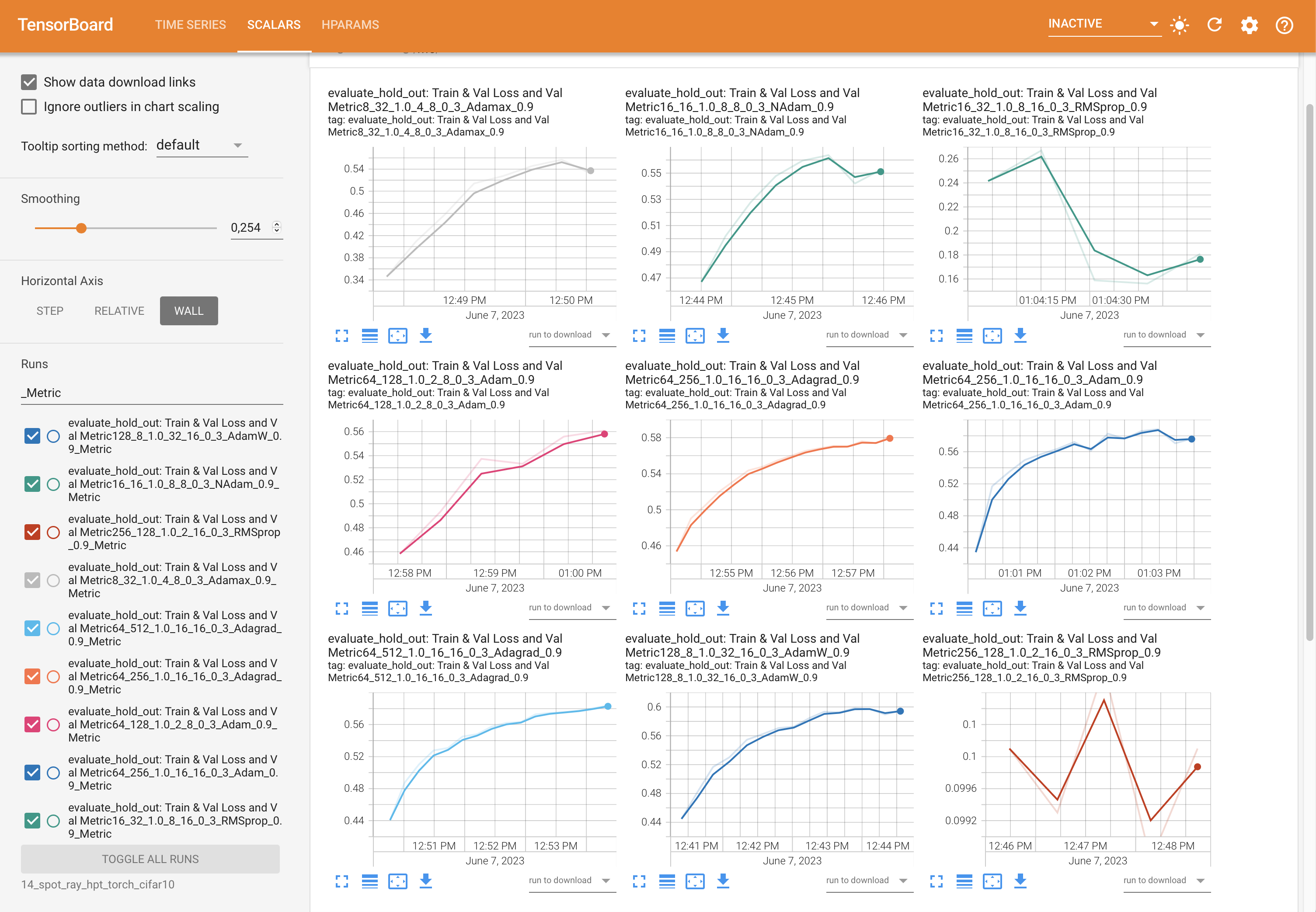}

}

\caption{\label{fig-tensorboard_0}Tensorboard}

\end{figure}

\begin{figure}

{\centering \includegraphics{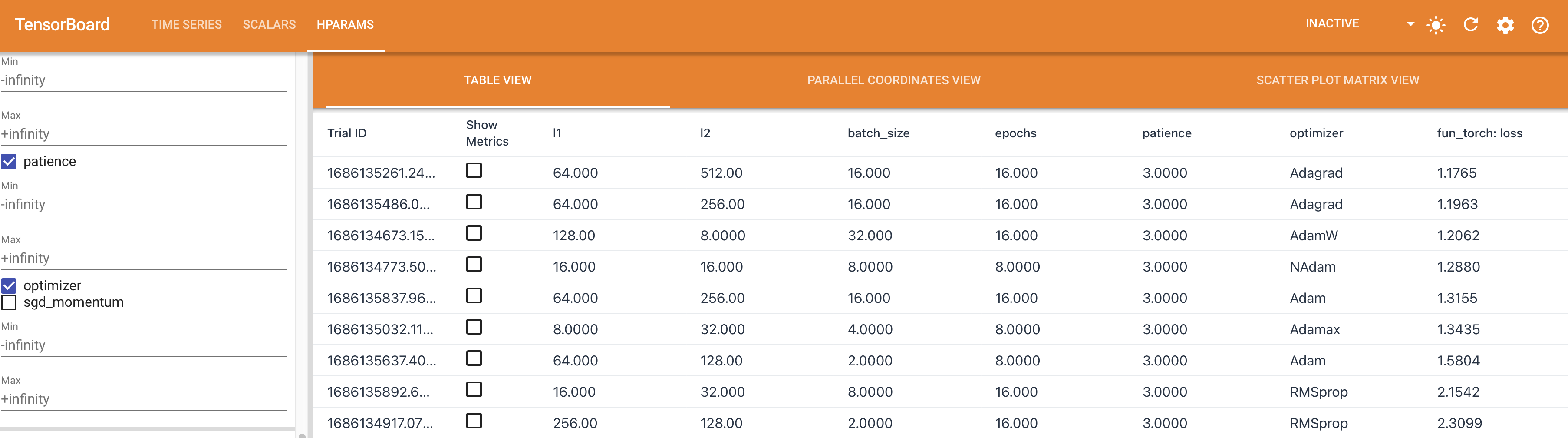}

}

\caption{\label{fig-tensorboard_hdparams}Tensorboard}

\end{figure}

\hypertarget{sec-results-tuning}{%
\subsection{Results}\label{sec-results-tuning}}

\begin{figure}

{\centering \includegraphics{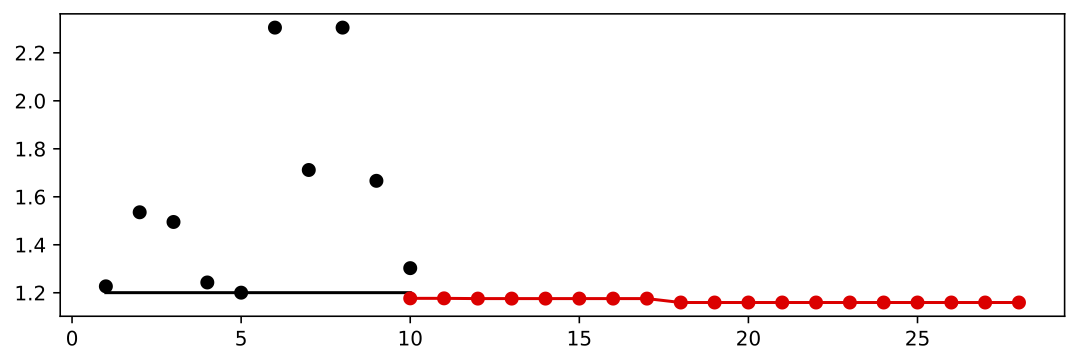}

}

\caption{\label{fig-progress}Progress plot. \texttt{Black} dots denote
results from the initial design. \texttt{Red} dots illustrate the
improvement found by the surrogate model based optimization (surrogate
model based optimization).}

\end{figure}

After the hyperparameter tuning run is finished, the progress of the
hyperparameter tuning can be visualized. The following code generates
the progress plot from Figure~\ref{fig-progress}.

\begin{Shaded}
\begin{Highlighting}[]
\NormalTok{spot\_tuner.plot\_progress(log\_y}\OperatorTok{=}\VariableTok{False}\NormalTok{, }
\NormalTok{    filename}\OperatorTok{=}\StringTok{"./figures/"} \OperatorTok{+}\NormalTok{ experiment\_name}\OperatorTok{+}\StringTok{"\_progress.png"}\NormalTok{)}
\end{Highlighting}
\end{Shaded}

Figure~\ref{fig-progress} shows a typical behaviour that can be observed
in many hyperparameter studies (Bartz et al. 2022): the largest
improvement is obtained during the evaluation of the initial design. The
surrogate model based optimization-optimization with the surrogate
refines the results. Figure~\ref{fig-progress} also illustrates one
major difference between \texttt{ray{[}tune{]}} as used in PyTorch
(2023a) and \texttt{spotPython}: the \texttt{ray{[}tune{]}} uses a
random search and will generate results similar to the \emph{black}
dots, whereas \texttt{spotPython} uses a surrogate model based
optimization and presents results represented by \emph{red} dots in
Figure~\ref{fig-progress}. The surrogate model based optimization is
considered to be more efficient than a random search, because the
surrogate model guides the search towards promising regions in the
hyperparameter space.

In addition to the improved (``optimized'') hyperparameter values,
\texttt{spotPython} allows a statistical analysis, e.g., a sensitivity
analysis, of the results. We can print the results of the hyperparameter
tuning, see Table~\ref{tbl-results}.

\begin{Shaded}
\begin{Highlighting}[]
\BuiltInTok{print}\NormalTok{(gen\_design\_table(fun\_control}\OperatorTok{=}\NormalTok{fun\_control, spot}\OperatorTok{=}\NormalTok{spot\_tuner))}
\end{Highlighting}
\end{Shaded}

\hypertarget{tbl-results}{}
\begin{longtable}[]{@{}
  >{\raggedright\arraybackslash}p{(\columnwidth - 16\tabcolsep) * \real{0.1071}}
  >{\raggedright\arraybackslash}p{(\columnwidth - 16\tabcolsep) * \real{0.0714}}
  >{\raggedright\arraybackslash}p{(\columnwidth - 16\tabcolsep) * \real{0.1607}}
  >{\raggedleft\arraybackslash}p{(\columnwidth - 16\tabcolsep) * \real{0.0893}}
  >{\raggedleft\arraybackslash}p{(\columnwidth - 16\tabcolsep) * \real{0.0893}}
  >{\raggedleft\arraybackslash}p{(\columnwidth - 16\tabcolsep) * \real{0.0893}}
  >{\raggedright\arraybackslash}p{(\columnwidth - 16\tabcolsep) * \real{0.1786}}
  >{\raggedleft\arraybackslash}p{(\columnwidth - 16\tabcolsep) * \real{0.1339}}
  >{\raggedright\arraybackslash}p{(\columnwidth - 16\tabcolsep) * \real{0.0804}}@{}}
\caption{\label{tbl-results}Results of the hyperparameter tuning. The
table shows the hyperparameters, their types, default values, lower and
upper bounds, and the transformation function. The column ``tuned''
shows the tuned values. The column ``importance'' shows the importance
of the hyperparameters. The column ``stars'' shows the importance of the
hyperparameters in stars. The importance is computed by the SPOT
software.}\tabularnewline
\toprule\noalign{}
\begin{minipage}[b]{\linewidth}\raggedright
name
\end{minipage} & \begin{minipage}[b]{\linewidth}\raggedright
type
\end{minipage} & \begin{minipage}[b]{\linewidth}\raggedright
default
\end{minipage} & \begin{minipage}[b]{\linewidth}\raggedleft
lower
\end{minipage} & \begin{minipage}[b]{\linewidth}\raggedleft
upper
\end{minipage} & \begin{minipage}[b]{\linewidth}\raggedleft
tuned
\end{minipage} & \begin{minipage}[b]{\linewidth}\raggedright
transform
\end{minipage} & \begin{minipage}[b]{\linewidth}\raggedleft
importance
\end{minipage} & \begin{minipage}[b]{\linewidth}\raggedright
stars
\end{minipage} \\
\midrule\noalign{}
\endfirsthead
\toprule\noalign{}
\begin{minipage}[b]{\linewidth}\raggedright
name
\end{minipage} & \begin{minipage}[b]{\linewidth}\raggedright
type
\end{minipage} & \begin{minipage}[b]{\linewidth}\raggedright
default
\end{minipage} & \begin{minipage}[b]{\linewidth}\raggedleft
lower
\end{minipage} & \begin{minipage}[b]{\linewidth}\raggedleft
upper
\end{minipage} & \begin{minipage}[b]{\linewidth}\raggedleft
tuned
\end{minipage} & \begin{minipage}[b]{\linewidth}\raggedright
transform
\end{minipage} & \begin{minipage}[b]{\linewidth}\raggedleft
importance
\end{minipage} & \begin{minipage}[b]{\linewidth}\raggedright
stars
\end{minipage} \\
\midrule\noalign{}
\endhead
\bottomrule\noalign{}
\endlastfoot
l1 & int & 5 & 2.0 & 9.0 & 7.0 & pow\_2\_int & 100.00 & *** \\
l2 & int & 5 & 2.0 & 9.0 & 3.0 & pow\_2\_int & 96.29 & *** \\
lr\_mult & float & 1.0 & 0.1 & 10.0 & 0.1 & None & 0.00 & \\
batchsize & int & 4 & 1.0 & 5.0 & 4.0 & pow\_2\_int & 0.00 & \\
epochs & int & 3 & 3.0 & 4.0 & 4.0 & pow\_2\_int & 4.18 & * \\
k\_folds & int & 2 & 0.0 & 0.0 & 0.0 & None & 0.00 & \\
patience & int & 5 & 3.0 & 3.0 & 3.0 & None & 0.00 & \\
optimizer & factor & SGD & 0.0 & 9.0 & 3.0 & None & 0.16 & . \\
\end{longtable}

\begin{figure}

{\centering \includegraphics[width=0.5\textwidth,height=\textheight]{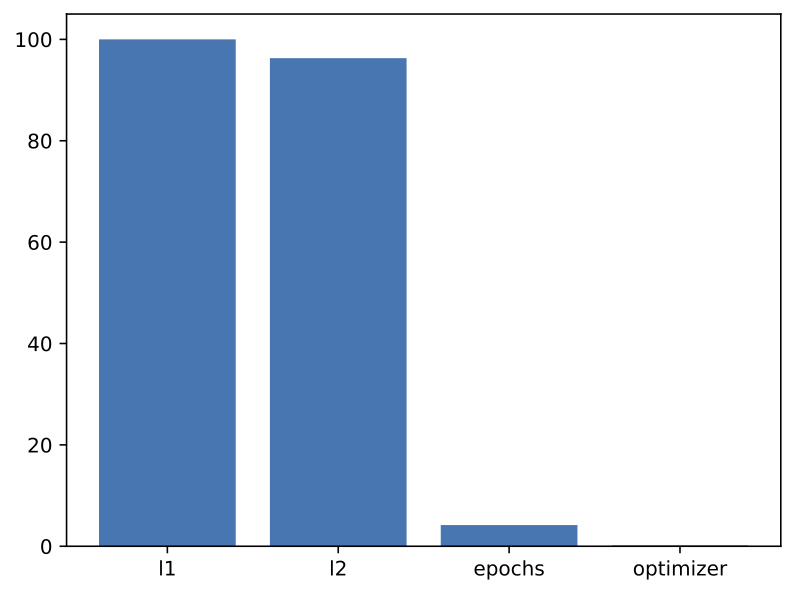}

}

\caption{\label{fig-importance}Variable importance}

\end{figure}

To visualize the most important hyperparameters, \texttt{spotPython}
provides the function \texttt{plot\_importance}. The following code
generates the importance plot from Figure~\ref{fig-importance}.

\begin{Shaded}
\begin{Highlighting}[]
\NormalTok{spot\_tuner.plot\_importance(threshold}\OperatorTok{=}\FloatTok{0.025}\NormalTok{,}
\NormalTok{    filename}\OperatorTok{=}\StringTok{"./figures/"} \OperatorTok{+}\NormalTok{ experiment\_name}\OperatorTok{+}\StringTok{"\_importance.png"}\NormalTok{)}
\end{Highlighting}
\end{Shaded}

\hypertarget{sec-get-spot-results}{%
\subsection{Get SPOT Results}\label{sec-get-spot-results}}

The architecture of the \texttt{spotPython} model can be obtained by the
following code:

\begin{Shaded}
\begin{Highlighting}[]
\ImportTok{from}\NormalTok{ spotPython.hyperparameters.values }\ImportTok{import}\NormalTok{ get\_one\_core\_model\_from\_X}
\NormalTok{X }\OperatorTok{=}\NormalTok{ spot\_tuner.to\_all\_dim(spot\_tuner.min\_X.reshape(}\DecValTok{1}\NormalTok{,}\OperatorTok{{-}}\DecValTok{1}\NormalTok{))}
\NormalTok{model\_spot }\OperatorTok{=}\NormalTok{ get\_one\_core\_model\_from\_X(X, fun\_control)}
\NormalTok{model\_spot}
\end{Highlighting}
\end{Shaded}

First, the numerical representation of the hyperparameters are obtained,
i.e., the numpy array \texttt{X} is generated. This array is then used
to generate the model \texttt{model\_spot} by the function
\texttt{get\_one\_core\_model\_from\_X}. The model \texttt{model\_spot}
has the following architecture:

\begin{Shaded}
\begin{Highlighting}[]
\NormalTok{Net\_CIFAR10(}
\NormalTok{  (conv1): Conv2d(3, 6, kernel\_size=(5, 5), stride=(1, 1))}
\NormalTok{  (pool): MaxPool2d(kernel\_size=2, stride=2, padding=0, dilation=1,}
\NormalTok{    ceil\_mode=False)}
\NormalTok{  (conv2): Conv2d(6, 16, kernel\_size=(5, 5), stride=(1, 1))}
\NormalTok{  (fc1): Linear(in\_features=400, out\_features=64, bias=True)}
\NormalTok{  (fc2): Linear(in\_features=64, out\_features=32, bias=True)}
\NormalTok{  (fc3): Linear(in\_features=32, out\_features=10, bias=True)}
\NormalTok{)}
\end{Highlighting}
\end{Shaded}

\hypertarget{get-default-hyperparameters}{%
\subsection{Get Default
Hyperparameters}\label{get-default-hyperparameters}}

In a similar manner as in Section~\ref{sec-get-spot-results}, the
default hyperparameters can be obtained.

\begin{Shaded}
\begin{Highlighting}[]
\CommentTok{\# fun\_control was modified, we generate a new one with the original }
\CommentTok{\# default hyperparameters}
\ImportTok{from}\NormalTok{ spotPython.hyperparameters.values }\ImportTok{import}\NormalTok{ get\_one\_core\_model\_from\_X}
\NormalTok{fc }\OperatorTok{=}\NormalTok{ fun\_control}
\NormalTok{fc.update(\{}\StringTok{"core\_model\_hyper\_dict"}\NormalTok{:}
\NormalTok{    hyper\_dict[fun\_control[}\StringTok{"core\_model"}\NormalTok{].}\VariableTok{\_\_name\_\_}\NormalTok{]\})}
\NormalTok{model\_default }\OperatorTok{=}\NormalTok{ get\_one\_core\_model\_from\_X(X\_start, fun\_control}\OperatorTok{=}\NormalTok{fc)}
\end{Highlighting}
\end{Shaded}

The corresponding default model has the following architecture:

\begin{Shaded}
\begin{Highlighting}[]
\NormalTok{Net\_CIFAR10(}
\NormalTok{  (conv1): Conv2d(3, 6, kernel\_size=(5, 5), stride=(1, 1))}
\NormalTok{  (pool): MaxPool2d(kernel\_size=2, stride=2, padding=0, dilation=1,}
\NormalTok{    ceil\_mode=False)}
\NormalTok{  (conv2): Conv2d(6, 16, kernel\_size=(5, 5), stride=(1, 1))}
\NormalTok{  (fc1): Linear(in\_features=400, out\_features=32, bias=True)}
\NormalTok{  (fc2): Linear(in\_features=32, out\_features=32, bias=True)}
\NormalTok{  (fc3): Linear(in\_features=32, out\_features=10, bias=True)}
\NormalTok{)}
\end{Highlighting}
\end{Shaded}

\hypertarget{evaluation-of-the-tuned-architecture}{%
\subsection{Evaluation of the Tuned
Architecture}\label{evaluation-of-the-tuned-architecture}}

The method \texttt{train\_tuned} takes a model architecture without
trained weights and trains this model with the train data. The train
data is split into train and validation data. The validation data is
used for early stopping. The trained model weights are saved as a
dictionary.

This evaluation is similar to the final evaluation in PyTorch (2023a).

\begin{Shaded}
\begin{Highlighting}[]
\ImportTok{from}\NormalTok{ spotPython.torch.traintest }\ImportTok{import}\NormalTok{ (}
\NormalTok{    train\_tuned,}
\NormalTok{    test\_tuned,}
\NormalTok{    )}
\NormalTok{train\_tuned(net}\OperatorTok{=}\NormalTok{model\_default, train\_dataset}\OperatorTok{=}\NormalTok{train, shuffle}\OperatorTok{=}\VariableTok{True}\NormalTok{,}
\NormalTok{        loss\_function}\OperatorTok{=}\NormalTok{fun\_control[}\StringTok{"loss\_function"}\NormalTok{],}
\NormalTok{        metric}\OperatorTok{=}\NormalTok{fun\_control[}\StringTok{"metric\_torch"}\NormalTok{],}
\NormalTok{        device }\OperatorTok{=}\NormalTok{ DEVICE, show\_batch\_interval}\OperatorTok{=}\DecValTok{1\_000\_000}\NormalTok{,}
\NormalTok{        path}\OperatorTok{=}\VariableTok{None}\NormalTok{,}
\NormalTok{        task}\OperatorTok{=}\NormalTok{fun\_control[}\StringTok{"task"}\NormalTok{],)}

\NormalTok{test\_tuned(net}\OperatorTok{=}\NormalTok{model\_default, test\_dataset}\OperatorTok{=}\NormalTok{test, }
\NormalTok{        loss\_function}\OperatorTok{=}\NormalTok{fun\_control[}\StringTok{"loss\_function"}\NormalTok{],}
\NormalTok{        metric}\OperatorTok{=}\NormalTok{fun\_control[}\StringTok{"metric\_torch"}\NormalTok{],}
\NormalTok{        shuffle}\OperatorTok{=}\VariableTok{False}\NormalTok{, }
\NormalTok{        device }\OperatorTok{=}\NormalTok{ DEVICE,}
\NormalTok{        task}\OperatorTok{=}\NormalTok{fun\_control[}\StringTok{"task"}\NormalTok{],)}
\end{Highlighting}
\end{Shaded}

The following code trains the model \texttt{model\_spot}. If
\texttt{path} is set to a filename, e.g.,
\texttt{path\ =\ "model\_spot\_trained.pt"}, the weights of the trained
model will be saved to this file.

\begin{Shaded}
\begin{Highlighting}[]
\NormalTok{train\_tuned(net}\OperatorTok{=}\NormalTok{model\_spot, train\_dataset}\OperatorTok{=}\NormalTok{train,}
\NormalTok{        loss\_function}\OperatorTok{=}\NormalTok{fun\_control[}\StringTok{"loss\_function"}\NormalTok{],}
\NormalTok{        metric}\OperatorTok{=}\NormalTok{fun\_control[}\StringTok{"metric\_torch"}\NormalTok{],}
\NormalTok{        shuffle}\OperatorTok{=}\VariableTok{True}\NormalTok{,}
\NormalTok{        device }\OperatorTok{=}\NormalTok{ DEVICE,}
\NormalTok{        path}\OperatorTok{=}\VariableTok{None}\NormalTok{,}
\NormalTok{        task}\OperatorTok{=}\NormalTok{fun\_control[}\StringTok{"task"}\NormalTok{],)}
\end{Highlighting}
\end{Shaded}

\begin{Shaded}
\begin{Highlighting}[]
\NormalTok{Loss on hold{-}out set: 1.2267619131326675}
\NormalTok{Accuracy on hold{-}out set: 0.58955}
\NormalTok{Early stopping at epoch 13}
\end{Highlighting}
\end{Shaded}

If \texttt{path} is set to a filename, e.g.,
\texttt{path\ =\ "model\_spot\_trained.pt"}, the weights of the trained
model will be loaded from this file.

\begin{Shaded}
\begin{Highlighting}[]
\NormalTok{test\_tuned(net}\OperatorTok{=}\NormalTok{model\_spot, test\_dataset}\OperatorTok{=}\NormalTok{test,}
\NormalTok{            shuffle}\OperatorTok{=}\VariableTok{False}\NormalTok{,}
\NormalTok{            loss\_function}\OperatorTok{=}\NormalTok{fun\_control[}\StringTok{"loss\_function"}\NormalTok{],}
\NormalTok{            metric}\OperatorTok{=}\NormalTok{fun\_control[}\StringTok{"metric\_torch"}\NormalTok{],}
\NormalTok{            device }\OperatorTok{=}\NormalTok{ DEVICE,}
\NormalTok{            task}\OperatorTok{=}\NormalTok{fun\_control[}\StringTok{"task"}\NormalTok{],)}
\end{Highlighting}
\end{Shaded}

\begin{Shaded}
\begin{Highlighting}[]
\NormalTok{Loss on hold{-}out set: 1.242568492603302}
\NormalTok{Accuracy on hold{-}out set: 0.5957}
\end{Highlighting}
\end{Shaded}

\hypertarget{comparison-with-default-hyperparameters-and-ray-tune}{%
\subsection{Comparison with Default Hyperparameters and Ray
Tune}\label{comparison-with-default-hyperparameters-and-ray-tune}}

Table~\ref{tbl-comparison} shows the loss and accuracy of the default
model, the model with the hyperparameters from SPOT, and the model with
the hyperparameters from \texttt{ray{[}tune{]}}.

\hypertarget{tbl-comparison}{}
\begin{longtable}[]{@{}
  >{\raggedright\arraybackslash}p{(\columnwidth - 8\tabcolsep) * \real{0.1618}}
  >{\raggedleft\arraybackslash}p{(\columnwidth - 8\tabcolsep) * \real{0.2500}}
  >{\raggedleft\arraybackslash}p{(\columnwidth - 8\tabcolsep) * \real{0.3088}}
  >{\raggedleft\arraybackslash}p{(\columnwidth - 8\tabcolsep) * \real{0.1324}}
  >{\raggedleft\arraybackslash}p{(\columnwidth - 8\tabcolsep) * \real{0.1471}}@{}}
\caption{\label{tbl-comparison}Comparison of the loss and accuracy of
the default model, the model with the hyperparameters from SPOT, and the
model with the hyperparameters from \texttt{ray{[}tune{]}}.
\texttt{ray{[}tune{]}} only shows the validation loss, because training
loss is not reported by \texttt{ray{[}tune{]}}.}\tabularnewline
\toprule\noalign{}
\begin{minipage}[b]{\linewidth}\raggedright
Model
\end{minipage} & \begin{minipage}[b]{\linewidth}\raggedleft
Validation Loss
\end{minipage} & \begin{minipage}[b]{\linewidth}\raggedleft
Validation Accuracy
\end{minipage} & \begin{minipage}[b]{\linewidth}\raggedleft
Loss
\end{minipage} & \begin{minipage}[b]{\linewidth}\raggedleft
Accuracy
\end{minipage} \\
\midrule\noalign{}
\endfirsthead
\toprule\noalign{}
\begin{minipage}[b]{\linewidth}\raggedright
Model
\end{minipage} & \begin{minipage}[b]{\linewidth}\raggedleft
Validation Loss
\end{minipage} & \begin{minipage}[b]{\linewidth}\raggedleft
Validation Accuracy
\end{minipage} & \begin{minipage}[b]{\linewidth}\raggedleft
Loss
\end{minipage} & \begin{minipage}[b]{\linewidth}\raggedleft
Accuracy
\end{minipage} \\
\midrule\noalign{}
\endhead
\bottomrule\noalign{}
\endlastfoot
Default & 2.1221 & 0.2452 & 2.1182 & 0.2425 \\
\texttt{spotPython} & 1.2268 & 0.5896 & 1.2426 & 0.5957 \\
\texttt{ray{[}tune{]}} & 1.1815 & 0.5836 & - & 0.5806 \\
\end{longtable}

\hypertarget{detailed-hyperparameter-plots}{%
\subsection{Detailed Hyperparameter
Plots}\label{detailed-hyperparameter-plots}}

The contour plots in this section visualize the interactions of the
three most important hyperparameters, \texttt{l1}, \texttt{l2}, and
\texttt{epochs}, and \texttt{optimizer} of the surrogate model used to
optimize the hyperparameters. Since some of these hyperparameters take
fatorial or integer values, sometimes step-like fitness landcapes (or
response surfaces) are generated. SPOT draws the interactions of the
main hyperparameters by default. It is also possible to visualize all
interactions. For this, again refer to the notebook (Bartz-Beielstein
2023).

\begin{Shaded}
\begin{Highlighting}[]
\NormalTok{filename }\OperatorTok{=} \StringTok{"./figures/"} \OperatorTok{+}\NormalTok{ experiment\_name}
\NormalTok{spot\_tuner.plot\_important\_hyperparameter\_contour(filename}\OperatorTok{=}\NormalTok{filename)}
\end{Highlighting}
\end{Shaded}

\begin{figure}

{\centering \includegraphics{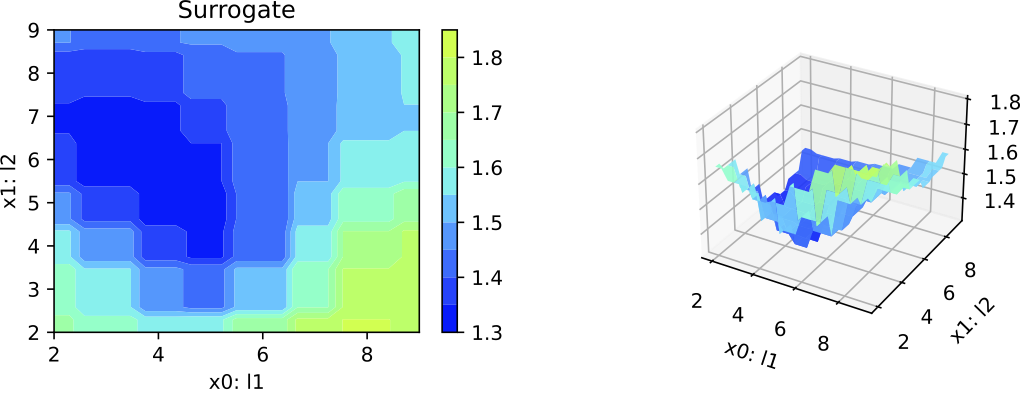}

}

\caption{\label{fig-contour-0-1}Contour plot of the loss as a function
of \texttt{l1} and \texttt{l2}, i.e., the number of neurons in the
layers.}

\end{figure}

\begin{figure}

{\centering \includegraphics{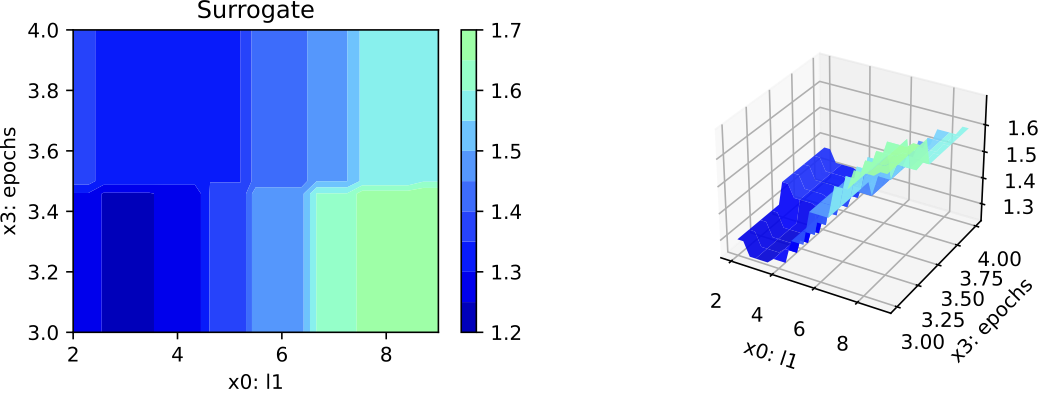}

}

\caption{\label{fig-contour-0-3}Contour plot of the loss as a function
of the number of epochs and the neurons in layer \texttt{l1}.}

\end{figure}

\begin{figure}

{\centering \includegraphics{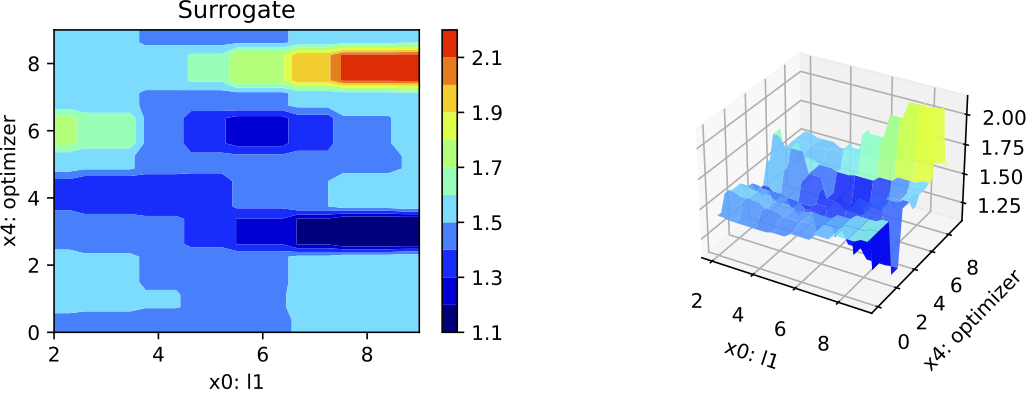}

}

\caption{\label{fig-contour-0-4}Contour plot of the loss as a function
of the optimizer and the neurons in layer \texttt{l1}.}

\end{figure}

\begin{figure}

{\centering \includegraphics{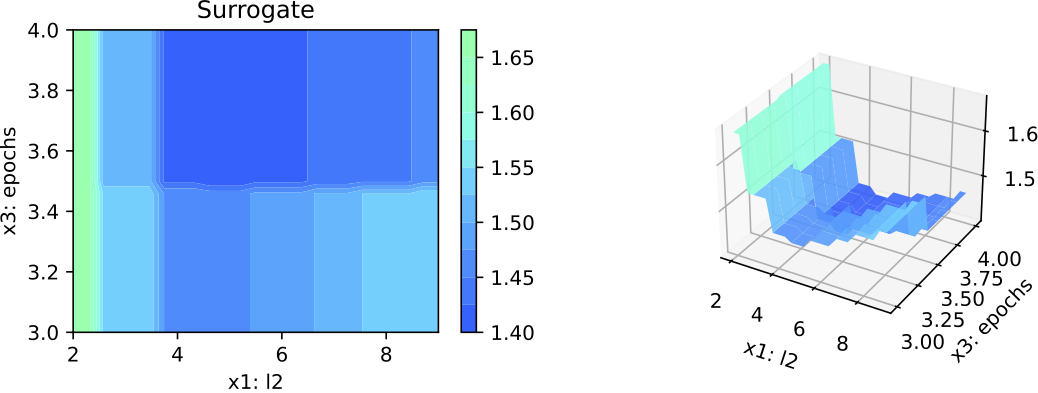}

}

\caption{\label{fig-contour-1-3}Contour plot of the loss as a function
of the number of epochs and the neurons in layer \texttt{l2}.}

\end{figure}

\begin{figure}

{\centering \includegraphics{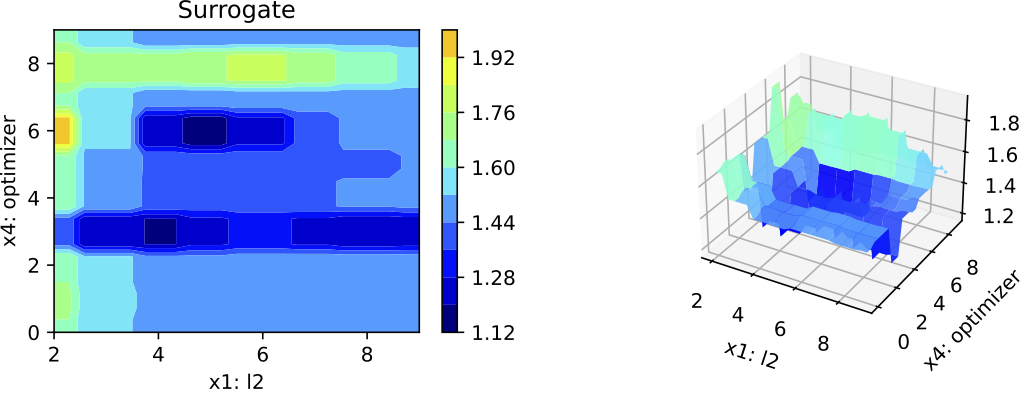}

}

\caption{\label{fig-contour-1-4}Contour plot of the loss as a function
of the optimizer and the neurons in layer \texttt{l2}.}

\end{figure}

\begin{figure}

{\centering \includegraphics{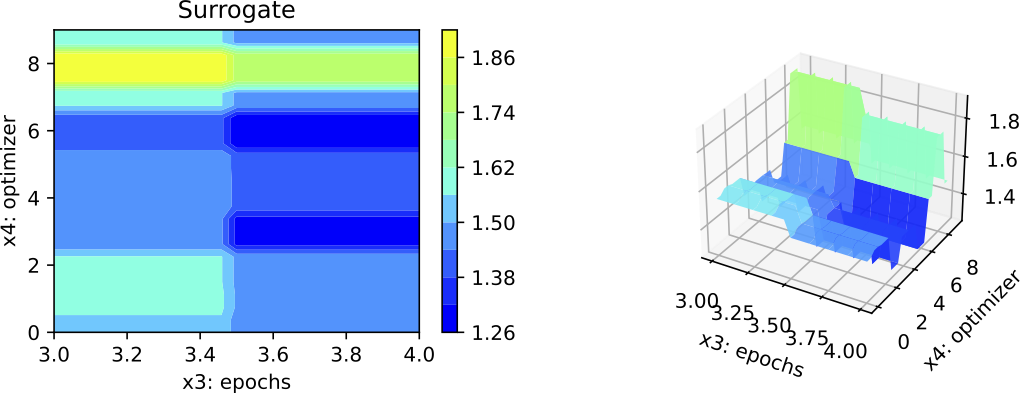}

}

\caption{\label{fig-contour-3-4}Contour plot of the loss as a function
of the optimizer and the number of epochs.}

\end{figure}

\begin{figure}

{\centering \includegraphics{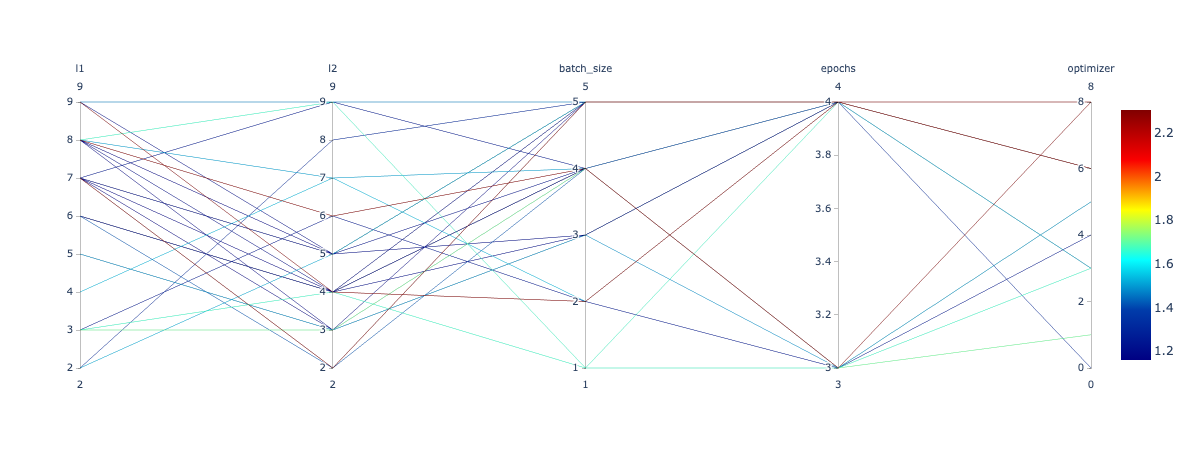}

}

\caption{\label{fig-parallel}Parallel plot}

\end{figure}

Figure~\ref{fig-contour-0-1} to Figure~\ref{fig-contour-3-4} show the
contour plots of the loss as a function of the hyperparameters. These
plots are very helpful for benchmark studies and for understanding
neural networks. \texttt{spotPython} provides additional tools for a
visual inspection of the results and give valuable insights into the
hyperparameter tuning process. This is especially useful for model
explainability, transparency, and trustworthiness. In addition to the
contour plots, Figure~\ref{fig-parallel} shows the parallel plot of the
hyperparameters.

\begin{Shaded}
\begin{Highlighting}[]
\NormalTok{spot\_tuner.parallel\_plot()}
\end{Highlighting}
\end{Shaded}

\hypertarget{sec-summary}{%
\section{Summary and Outlook}\label{sec-summary}}

This tutorial presents the hyperparameter tuning open source software
\texttt{spotPython} for \texttt{PyTorch}. To show its basic features, a
comparison with the ``official'' \texttt{PyTorch} hyperparameter tuning
tutorial (PyTorch 2023a) is presented. Some of the advantages of
\texttt{spotPython} are:

\begin{itemize}
\tightlist
\item
  Numerical and categorical hyperparameters.
\item
  Powerful surrogate models.
\item
  Flexible approach and easy to use.
\item
  Simple JSON files for the specification of the hyperparameters.
\item
  Extension of default and user specified network classes.
\item
  Noise handling techniques.
\end{itemize}

Currently, only rudimentary parallel and distributed neural network
training is possible, but these capabilities will be extended in the
future. The next version of \texttt{spotPython} will also include a more
detailed documentation and more examples.

\begin{tcolorbox}[enhanced jigsaw, colbacktitle=quarto-callout-important-color!10!white, coltitle=black, opacitybacktitle=0.6, title=\textcolor{quarto-callout-important-color}{\faExclamation}\hspace{0.5em}{Important}, bottomrule=.15mm, arc=.35mm, leftrule=.75mm, breakable, colframe=quarto-callout-important-color-frame, colback=white, left=2mm, toptitle=1mm, bottomtitle=1mm, rightrule=.15mm, titlerule=0mm, toprule=.15mm, opacityback=0]

Important: This tutorial does not present a complete benchmarking study
(Bartz-Beielstein et al. 2020). The results are only preliminary and
highly dependent on the local configuration (hard- and software). Our
goal is to provide a first impression of the performance of the
hyperparameter tuning package \texttt{spotPython}. To demonstrate its
capabilities, a quick comparison with \texttt{ray{[}tune{]}} was
performed. \texttt{ray{[}tune{]}} was chosen, because it is presented as
``an industry standard tool for distributed hyperparameter tuning.'' The
results should be interpreted with care.

\end{tcolorbox}

\hypertarget{sec-appendix}{%
\section{Appendix}\label{sec-appendix}}

\hypertarget{sample-output-from-ray-tunes-run}{%
\subsection{Sample Output From Ray Tune's
Run}\label{sample-output-from-ray-tunes-run}}

The output from \texttt{ray{[}tune{]}} could look like this (PyTorch
2023b):

\begin{Shaded}
\begin{Highlighting}[]
\NormalTok{Number of trials: 10 (10 TERMINATED)}
\NormalTok{{-}{-}{-}{-}{-}{-}+{-}{-}{-}{-}{-}{-}+{-}{-}{-}{-}{-}{-}{-}{-}{-}{-}{-}{-}{-}+{-}{-}{-}{-}{-}{-}{-}{-}{-}{-}{-}{-}{-}{-}+{-}{-}{-}{-}{-}{-}{-}{-}{-}+{-}{-}{-}{-}{-}{-}{-}{-}{-}{-}{-}{-}+{-}{-}{-}{-}{-}{-}{-}{-}{-}{-}{-}{-}{-}{-}{-}{-}{-}{-}{-}{-}+}
\NormalTok{|   l1 |   l2 |          lr |   batch\_size |    loss |   accuracy | training\_iteration |}
\NormalTok{+{-}{-}{-}{-}{-}{-}+{-}{-}{-}{-}{-}{-}+{-}{-}{-}{-}{-}{-}{-}{-}{-}{-}{-}{-}{-}+{-}{-}{-}{-}{-}{-}{-}{-}{-}{-}{-}{-}{-}{-}+{-}{-}{-}{-}{-}{-}{-}{-}{-}+{-}{-}{-}{-}{-}{-}{-}{-}{-}{-}{-}{-}+{-}{-}{-}{-}{-}{-}{-}{-}{-}{-}{-}{-}{-}{-}{-}{-}{-}{-}{-}{-}|}
\NormalTok{|   64 |    4 | 0.00011629  |            2 | 1.87273 |     0.244  |                  2 |}
\NormalTok{|   32 |   64 | 0.000339763 |            8 | 1.23603 |     0.567  |                  8 |}
\NormalTok{|    8 |   16 | 0.00276249  |           16 | 1.1815  |     0.5836 |                 10 |}
\NormalTok{|    4 |   64 | 0.000648721 |            4 | 1.31131 |     0.5224 |                  8 |}
\NormalTok{|   32 |   16 | 0.000340753 |            8 | 1.26454 |     0.5444 |                  8 |}
\NormalTok{|    8 |    4 | 0.000699775 |            8 | 1.99594 |     0.1983 |                  2 |}
\NormalTok{|  256 |    8 | 0.0839654   |           16 | 2.3119  |     0.0993 |                  1 |}
\NormalTok{|   16 |  128 | 0.0758154   |           16 | 2.33575 |     0.1327 |                  1 |}
\NormalTok{|   16 |    8 | 0.0763312   |           16 | 2.31129 |     0.1042 |                  4 |}
\NormalTok{|  128 |   16 | 0.000124903 |            4 | 2.26917 |     0.1945 |                  1 |}
\NormalTok{+{-}{-}{-}{-}{-}+{-}{-}{-}{-}{-}{-}+{-}{-}{-}{-}{-}{-}+{-}{-}{-}{-}{-}{-}{-}{-}{-}{-}{-}{-}{-}+{-}{-}{-}{-}{-}{-}{-}{-}{-}{-}{-}{-}{-}{-}+{-}{-}{-}{-}{-}{-}{-}{-}{-}+{-}{-}{-}{-}{-}{-}{-}{-}{-}{-}{-}{-}+{-}{-}{-}{-}{-}{-}{-}{-}{-}{-}{-}{-}{-}{-}{-}{-}{-}{-}{-}{-}+}
\NormalTok{Best trial config: \{\textquotesingle{}l1\textquotesingle{}: 8, \textquotesingle{}l2\textquotesingle{}: 16, \textquotesingle{}lr\textquotesingle{}: 0.00276249, \textquotesingle{}batch\_size\textquotesingle{}: 16, \textquotesingle{}data\_dir\textquotesingle{}: \textquotesingle{}...\textquotesingle{}\}}
\NormalTok{Best trial final validation loss: 1.181501}
\NormalTok{Best trial final validation accuracy: 0.5836}
\NormalTok{Best trial test set accuracy: 0.5806}
\end{Highlighting}
\end{Shaded}

\newpage{}

\hypertarget{sec-data-splitting}{%
\subsection{Detailed Description of the Data
Splitting}\label{sec-data-splitting}}

\hypertarget{description-of-the-train_hold_out-setting}{%
\subsubsection{\texorpdfstring{Description of the
\texttt{"train\_hold\_out"}
Setting}{Description of the "train\_hold\_out" Setting}}\label{description-of-the-train_hold_out-setting}}

The \texttt{"train\_hold\_out"} setting is used by default. It uses the
loss function specfied in \texttt{fun\_control} and the metric specified
in \texttt{fun\_control}.

\begin{enumerate}
\def\labelenumi{\arabic{enumi}.}
\tightlist
\item
  First, the method \texttt{HyperTorch().fun\_torch} is called.
\item
  \texttt{fun\_torc()}, which is implemented in the file
  \texttt{hypertorch.py}, calls \texttt{evaluate\_hold\_out()} as
  follows:
\end{enumerate}

\begin{Shaded}
\begin{Highlighting}[]
\NormalTok{df\_eval, \_ = evaluate\_hold\_out(}
\NormalTok{    model,}
\NormalTok{    train\_dataset=fun\_control["train"],}
\NormalTok{    shuffle=self.fun\_control["shuffle"],}
\NormalTok{    loss\_function=self.fun\_control["loss\_function"],}
\NormalTok{    metric=self.fun\_control["metric\_torch"],}
\NormalTok{    device=self.fun\_control["device"],}
\NormalTok{    show\_batch\_interval=self.fun\_control["show\_batch\_interval"],}
\NormalTok{    path=self.fun\_control["path"],}
\NormalTok{    task=self.fun\_control["task"],}
\NormalTok{    writer=self.fun\_control["writer"],}
\NormalTok{    writerId=config\_id,}
\NormalTok{)}
\end{Highlighting}
\end{Shaded}

Note: Only the data set \texttt{fun\_control{[}"train"{]}} is used for
training and validation. It is used in \texttt{evaluate\_hold\_out} as
follows:

\begin{Shaded}
\begin{Highlighting}[]
\NormalTok{trainloader, valloader = create\_train\_val\_data\_loaders(}
\NormalTok{                dataset=train\_dataset, batch\_size=batch\_size\_instance, shuffle=shuffle}
\NormalTok{            )}
\end{Highlighting}
\end{Shaded}

\texttt{create\_train\_val\_data\_loaders()} splits the
\texttt{train\_dataset} into \texttt{trainloader} and \texttt{valloader}
using \texttt{torch.utils.data.random\_split()} as follows:

\begin{Shaded}
\begin{Highlighting}[]
\NormalTok{def create\_train\_val\_data\_loaders(dataset, batch\_size, shuffle, num\_workers=0):}
\NormalTok{    test\_abs = int(len(dataset) * 0.6)}
\NormalTok{    train\_subset, val\_subset = random\_split(dataset, [test\_abs, len(dataset) {-} test\_abs])}
\NormalTok{    trainloader = torch.utils.data.DataLoader(}
\NormalTok{        train\_subset, batch\_size=int(batch\_size), shuffle=shuffle, num\_workers=num\_workers}
\NormalTok{    )}
\NormalTok{    valloader = torch.utils.data.DataLoader(}
\NormalTok{        val\_subset, batch\_size=int(batch\_size), shuffle=shuffle, num\_workers=num\_workers}
\NormalTok{    )}
\NormalTok{    return trainloader, valloader}
\end{Highlighting}
\end{Shaded}

The optimizer is set up as follows:

\begin{Shaded}
\begin{Highlighting}[]
\NormalTok{optimizer\_instance = net.optimizer}
\NormalTok{lr\_mult\_instance = net.lr\_mult}
\NormalTok{sgd\_momentum\_instance = net.sgd\_momentum}
\NormalTok{optimizer = optimizer\_handler(}
\NormalTok{    optimizer\_name=optimizer\_instance,}
\NormalTok{    params=net.parameters(),}
\NormalTok{    lr\_mult=lr\_mult\_instance,}
\NormalTok{    sgd\_momentum=sgd\_momentum\_instance,}
\NormalTok{)}
\end{Highlighting}
\end{Shaded}

\begin{enumerate}
\def\labelenumi{\arabic{enumi}.}
\setcounter{enumi}{2}
\tightlist
\item
  \texttt{evaluate\_hold\_out()} sets the \texttt{net} attributes such
  as \texttt{epochs}, \texttt{batch\_size}, \texttt{optimizer}, and
  \texttt{patience}. For each epoch, the methods
  \texttt{train\_one\_epoch()} and \texttt{validate\_one\_epoch()} are
  called, the former for training and the latter for validation and
  early stopping. The validation loss from the last epoch (not the best
  validation loss) is returned from \texttt{evaluate\_hold\_out}.
\item
  The method \texttt{train\_one\_epoch()} is implemented as follows:
\end{enumerate}

\begin{Shaded}
\begin{Highlighting}[]
\NormalTok{def train\_one\_epoch(}
\NormalTok{    net,}
\NormalTok{    trainloader,}
\NormalTok{    batch\_size,}
\NormalTok{    loss\_function,}
\NormalTok{    optimizer,}
\NormalTok{    device,}
\NormalTok{    show\_batch\_interval=10\_000,}
\NormalTok{    task=None,}
\NormalTok{):}
\NormalTok{    running\_loss = 0.0}
\NormalTok{    epoch\_steps = 0}
\NormalTok{    for batch\_nr, data in enumerate(trainloader, 0):}
\NormalTok{        input, target = data}
\NormalTok{        input, target = input.to(device), target.to(device)}
\NormalTok{        optimizer.zero\_grad()}
\NormalTok{        output = net(input)}
\NormalTok{        if task == "regression":}
\NormalTok{            target = target.unsqueeze(1)}
\NormalTok{            if target.shape == output.shape:}
\NormalTok{                loss = loss\_function(output, target)}
\NormalTok{            else:}
\NormalTok{                raise ValueError(f"Shapes of target and output do not match:}
\NormalTok{                 \{target.shape\} vs \{output.shape\}")}
\NormalTok{        elif task == "classification":}
\NormalTok{            loss = loss\_function(output, target)}
\NormalTok{        else:}
\NormalTok{            raise ValueError(f"Unknown task: \{task\}")}
\NormalTok{        loss.backward()}
\NormalTok{        torch.nn.utils.clip\_grad\_norm\_(net.parameters(), max\_norm=1.0)}
\NormalTok{        optimizer.step()}
\NormalTok{        running\_loss += loss.item()}
\NormalTok{        epoch\_steps += 1}
\NormalTok{        if batch\_nr \% show\_batch\_interval == (show\_batch\_interval {-} 1):  }
\NormalTok{            print(}
\NormalTok{                "Batch: \%5d. Batch Size: \%d. Training Loss (running): \%.3f"}
\NormalTok{                \% (batch\_nr + 1, int(batch\_size), running\_loss / epoch\_steps)}
\NormalTok{            )}
\NormalTok{            running\_loss = 0.0}
\NormalTok{    return loss.item()}
\end{Highlighting}
\end{Shaded}

\begin{enumerate}
\def\labelenumi{\arabic{enumi}.}
\setcounter{enumi}{4}
\tightlist
\item
  The method \texttt{validate\_one\_epoch()} is implemented as follows:
\end{enumerate}

\begin{Shaded}
\begin{Highlighting}[]
\NormalTok{def validate\_one\_epoch(net, valloader, loss\_function, metric, device, task):}
\NormalTok{    val\_loss = 0.0}
\NormalTok{    val\_steps = 0}
\NormalTok{    total = 0}
\NormalTok{    correct = 0}
\NormalTok{    metric.reset()}
\NormalTok{    for i, data in enumerate(valloader, 0):}
\NormalTok{        \# get batches}
\NormalTok{        with torch.no\_grad():}
\NormalTok{            input, target = data}
\NormalTok{            input, target = input.to(device), target.to(device)}
\NormalTok{            output = net(input)}
\NormalTok{            \# print(f"target: \{target\}")}
\NormalTok{            \# print(f"output: \{output\}")}
\NormalTok{            if task == "regression":}
\NormalTok{                target = target.unsqueeze(1)}
\NormalTok{                if target.shape == output.shape:}
\NormalTok{                    loss = loss\_function(output, target)}
\NormalTok{                else:}
\NormalTok{                    raise ValueError(f"Shapes of target and output }
\NormalTok{                        do not match: \{target.shape\} vs \{output.shape\}")}
\NormalTok{                metric\_value = metric.update(output, target)}
\NormalTok{            elif task == "classification":}
\NormalTok{                loss = loss\_function(output, target)}
\NormalTok{                metric\_value = metric.update(output, target)}
\NormalTok{                \_, predicted = torch.max(output.data, 1)}
\NormalTok{                total += target.size(0)}
\NormalTok{                correct += (predicted == target).sum().item()}
\NormalTok{            else:}
\NormalTok{                raise ValueError(f"Unknown task: \{task\}")}
\NormalTok{            val\_loss += loss.cpu().numpy()}
\NormalTok{            val\_steps += 1}
\NormalTok{    loss = val\_loss / val\_steps}
\NormalTok{    print(f"Loss on hold{-}out set: \{loss\}")}
\NormalTok{    if task == "classification":}
\NormalTok{        accuracy = correct / total}
\NormalTok{        print(f"Accuracy on hold{-}out set: \{accuracy\}")}
\NormalTok{    \# metric on all batches using custom accumulation}
\NormalTok{    metric\_value = metric.compute()}
\NormalTok{    metric\_name = type(metric).\_\_name\_\_}
\NormalTok{    print(f"\{metric\_name\} value on hold{-}out data: \{metric\_value\}")}
\NormalTok{    return metric\_value, loss}
\end{Highlighting}
\end{Shaded}

\hypertarget{description-of-the-test_hold_out-setting}{%
\subsubsection{\texorpdfstring{Description of the
\texttt{"test\_hold\_out"}
Setting}{Description of the "test\_hold\_out" Setting}}\label{description-of-the-test_hold_out-setting}}

It uses the loss function specfied in \texttt{fun\_control} and the
metric specified in \texttt{fun\_control}.

\begin{enumerate}
\def\labelenumi{\arabic{enumi}.}
\tightlist
\item
  First, the method \texttt{HyperTorch().fun\_torch} is called.
\item
  \texttt{fun\_torc()} calls
  \texttt{spotPython.torch.traintest.evaluate\_hold\_out()} similar to
  the \texttt{"train\_hold\_out"} setting with one exception: It passes
  an additional \texttt{test} data set to \texttt{evaluate\_hold\_out()}
  as follows:
\end{enumerate}

\begin{Shaded}
\begin{Highlighting}[]
\NormalTok{test\_dataset=fun\_control["test"]}
\end{Highlighting}
\end{Shaded}

\texttt{evaluate\_hold\_out()} calls
\texttt{create\_train\_test\_data\_loaders} instead of
\texttt{create\_train\_val\_data\_loaders}: The two data sets are used
in \texttt{create\_train\_test\_data\_loaders} as follows:

\begin{Shaded}
\begin{Highlighting}[]
\NormalTok{def create\_train\_test\_data\_loaders(dataset, batch\_size, shuffle, test\_dataset, }
\NormalTok{        num\_workers=0):}
\NormalTok{    trainloader = torch.utils.data.DataLoader(}
\NormalTok{        dataset, batch\_size=int(batch\_size), shuffle=shuffle, }
\NormalTok{        num\_workers=num\_workers}
\NormalTok{    )}
\NormalTok{    testloader = torch.utils.data.DataLoader(}
\NormalTok{        test\_dataset, batch\_size=int(batch\_size), shuffle=shuffle, }
\NormalTok{        num\_workers=num\_workers}
\NormalTok{    )}
\NormalTok{    return trainloader, testloader}
\end{Highlighting}
\end{Shaded}

\begin{enumerate}
\def\labelenumi{\arabic{enumi}.}
\setcounter{enumi}{2}
\tightlist
\item
  The following steps are identical to the \texttt{"train\_hold\_out"}
  setting. Only a different data loader is used for testing.
\end{enumerate}

\hypertarget{detailed-description-of-the-train_cv-setting}{%
\subsubsection{\texorpdfstring{Detailed Description of the
\texttt{"train\_cv"}
Setting}{Detailed Description of the "train\_cv" Setting}}\label{detailed-description-of-the-train_cv-setting}}

It uses the loss function specfied in \texttt{fun\_control} and the
metric specified in \texttt{fun\_control}.

\begin{enumerate}
\def\labelenumi{\arabic{enumi}.}
\tightlist
\item
  First, the method \texttt{HyperTorch().fun\_torch} is called.
\item
  \texttt{fun\_torc()} calls
  \texttt{spotPython.torch.traintest.evaluate\_cv()} as follows (Note:
  Only the data set \texttt{fun\_control{[}"train"{]}} is used for CV.):
\end{enumerate}

\begin{Shaded}
\begin{Highlighting}[]
\NormalTok{df\_eval, \_ = evaluate\_cv(}
\NormalTok{    model,}
\NormalTok{    dataset=fun\_control["train"],}
\NormalTok{    shuffle=self.fun\_control["shuffle"],}
\NormalTok{    device=self.fun\_control["device"],}
\NormalTok{    show\_batch\_interval=self.fun\_control["show\_batch\_interval"],}
\NormalTok{    task=self.fun\_control["task"],}
\NormalTok{    writer=self.fun\_control["writer"],}
\NormalTok{    writerId=config\_id,}
\NormalTok{)}
\end{Highlighting}
\end{Shaded}

\begin{enumerate}
\def\labelenumi{\arabic{enumi}.}
\setcounter{enumi}{2}
\tightlist
\item
  In `evaluate\_cv(), the following steps are performed: The optimizer
  is set up as follows:
\end{enumerate}

\begin{Shaded}
\begin{Highlighting}[]
\NormalTok{optimizer\_instance = net.optimizer}
\NormalTok{lr\_instance = net.lr}
\NormalTok{sgd\_momentum\_instance = net.sgd\_momentum}
\NormalTok{optimizer = optimizer\_handler(optimizer\_name=optimizer\_instance,}
\NormalTok{     params=net.parameters(), lr\_mult=lr\_mult\_instance)}
\end{Highlighting}
\end{Shaded}

\texttt{evaluate\_cv()} sets the \texttt{net} attributes such as
\texttt{epochs}, \texttt{batch\_size}, \texttt{optimizer}, and
\texttt{patience}. CV is implemented as follows:

\begin{Shaded}
\begin{Highlighting}[]
\NormalTok{def evaluate\_cv(}
\NormalTok{    net,}
\NormalTok{    dataset,}
\NormalTok{    shuffle=False,}
\NormalTok{    loss\_function=None,}
\NormalTok{    num\_workers=0,}
\NormalTok{    device=None,}
\NormalTok{    show\_batch\_interval=10\_000,}
\NormalTok{    metric=None,}
\NormalTok{    path=None,}
\NormalTok{    task=None,}
\NormalTok{    writer=None,}
\NormalTok{    writerId=None,}
\NormalTok{):}
\NormalTok{    lr\_mult\_instance = net.lr\_mult}
\NormalTok{    epochs\_instance = net.epochs}
\NormalTok{    batch\_size\_instance = net.batch\_size}
\NormalTok{    k\_folds\_instance = net.k\_folds}
\NormalTok{    optimizer\_instance = net.optimizer}
\NormalTok{    patience\_instance = net.patience}
\NormalTok{    sgd\_momentum\_instance = net.sgd\_momentum}
\NormalTok{    removed\_attributes, net = get\_removed\_attributes\_and\_base\_net(net)}
\NormalTok{    metric\_values = \{\}}
\NormalTok{    loss\_values = \{\}}
\NormalTok{    try:}
\NormalTok{        device = getDevice(device=device)}
\NormalTok{        if torch.cuda.is\_available():}
\NormalTok{            device = "cuda:0"}
\NormalTok{            if torch.cuda.device\_count() \textgreater{} 1:}
\NormalTok{                print("We will use", torch.cuda.device\_count(), "GPUs!")}
\NormalTok{                net = nn.DataParallel(net)}
\NormalTok{        net.to(device)}
\NormalTok{        optimizer = optimizer\_handler(}
\NormalTok{            optimizer\_name=optimizer\_instance,}
\NormalTok{            params=net.parameters(),}
\NormalTok{            lr\_mult=lr\_mult\_instance,}
\NormalTok{            sgd\_momentum=sgd\_momentum\_instance,}
\NormalTok{        )}
\NormalTok{        kfold = KFold(n\_splits=k\_folds\_instance, shuffle=shuffle)}
\NormalTok{        for fold, (train\_ids, val\_ids) in enumerate(kfold.split(dataset)):}
\NormalTok{            print(f"Fold: \{fold + 1\}")}
\NormalTok{            train\_subsampler = torch.utils.data.SubsetRandomSampler(train\_ids)}
\NormalTok{            val\_subsampler = torch.utils.data.SubsetRandomSampler(val\_ids)}
\NormalTok{            trainloader = torch.utils.data.DataLoader(}
\NormalTok{                dataset, batch\_size=batch\_size\_instance, }
\NormalTok{                sampler=train\_subsampler, num\_workers=num\_workers}
\NormalTok{            )}
\NormalTok{            valloader = torch.utils.data.DataLoader(}
\NormalTok{                dataset, batch\_size=batch\_size\_instance, }
\NormalTok{                sampler=val\_subsampler, num\_workers=num\_workers}
\NormalTok{            )}
\NormalTok{            \# each fold starts with new weights:}
\NormalTok{            reset\_weights(net)}
\NormalTok{            \# Early stopping parameters}
\NormalTok{            best\_val\_loss = float("inf")}
\NormalTok{            counter = 0}
\NormalTok{            for epoch in range(epochs\_instance):}
\NormalTok{                print(f"Epoch: \{epoch + 1\}")}
\NormalTok{                \# training loss from one epoch:}
\NormalTok{                training\_loss = train\_one\_epoch(}
\NormalTok{                    net=net,}
\NormalTok{                    trainloader=trainloader,}
\NormalTok{                    batch\_size=batch\_size\_instance,}
\NormalTok{                    loss\_function=loss\_function,}
\NormalTok{                    optimizer=optimizer,}
\NormalTok{                    device=device,}
\NormalTok{                    show\_batch\_interval=show\_batch\_interval,}
\NormalTok{                    task=task,}
\NormalTok{                )}
\NormalTok{                \# Early stopping check. Calculate validation loss from one epoch:}
\NormalTok{                metric\_values[fold], loss\_values[fold] = validate\_one\_epoch(}
\NormalTok{                    net, valloader=valloader, loss\_function=loss\_function, }
\NormalTok{                    metric=metric, device=device, task=task}
\NormalTok{                )}
\NormalTok{                \# Log the running loss averaged per batch}
\NormalTok{                metric\_name = "Metric"}
\NormalTok{                if metric is None:}
\NormalTok{                    metric\_name = type(metric).\_\_name\_\_}
\NormalTok{                    print(f"\{metric\_name\} value on hold{-}out data: }
\NormalTok{                        \{metric\_values[fold]\}")}
\NormalTok{                if writer is not None:}
\NormalTok{                    writer.add\_scalars(}
\NormalTok{                        "evaluate\_cv fold:" + str(fold + 1) + }
\NormalTok{                        ". Train \& Val Loss and Val Metric" + writerId,}
\NormalTok{                        \{"Train loss": training\_loss, "Val loss": }
\NormalTok{                        loss\_values[fold], metric\_name: metric\_values[fold]\},}
\NormalTok{                        epoch + 1,}
\NormalTok{                    )}
\NormalTok{                    writer.flush()}
\NormalTok{                if loss\_values[fold] \textless{} best\_val\_loss:}
\NormalTok{                    best\_val\_loss = loss\_values[fold]}
\NormalTok{                    counter = 0}
\NormalTok{                    \# save model:}
\NormalTok{                    if path is not None:}
\NormalTok{                        torch.save(net.state\_dict(), path)}
\NormalTok{                else:}
\NormalTok{                    counter += 1}
\NormalTok{                    if counter \textgreater{}= patience\_instance:}
\NormalTok{                        print(f"Early stopping at epoch \{epoch\}")}
\NormalTok{                        break}
\NormalTok{        df\_eval = sum(loss\_values.values()) / len(loss\_values.values())}
\NormalTok{        df\_metrics = sum(metric\_values.values()) / len(metric\_values.values())}
\NormalTok{        df\_preds = np.nan}
\NormalTok{    except Exception as err:}
\NormalTok{        print(f"Error in Net\_Core. Call to evaluate\_cv() failed. \{err=\}, }
\NormalTok{            \{type(err)=\}")}
\NormalTok{        df\_eval = np.nan}
\NormalTok{        df\_preds = np.nan}
\NormalTok{    add\_attributes(net, removed\_attributes)}
\NormalTok{    if writer is not None:}
\NormalTok{        metric\_name = "Metric"}
\NormalTok{        if metric is None:}
\NormalTok{            metric\_name = type(metric).\_\_name\_\_}
\NormalTok{        writer.add\_scalars(}
\NormalTok{            "CV: Val Loss and Val Metric" + writerId,}
\NormalTok{            \{"CV{-}loss": df\_eval, metric\_name: df\_metrics\},}
\NormalTok{            epoch + 1,}
\NormalTok{        )}
\NormalTok{        writer.flush()}
\NormalTok{    return df\_eval, df\_preds, df\_metrics}
\end{Highlighting}
\end{Shaded}

\begin{enumerate}
\def\labelenumi{\arabic{enumi}.}
\setcounter{enumi}{3}
\item
  The method \texttt{train\_fold()} is implemented as shown above.
\item
  The method \texttt{validate\_one\_epoch()} is implemented as shown
  above. In contrast to the hold-out setting, it is called for each of
  the \(k\) folds. The results are stored in a dictionaries
  \texttt{metric\_values} and \texttt{loss\_values}. The results are
  averaged over the \(k\) folds and returned as \texttt{df\_eval}.
\end{enumerate}

\hypertarget{detailed-description-of-the-test_cv-setting}{%
\subsubsection{\texorpdfstring{Detailed Description of the
\texttt{"test\_cv"}
Setting}{Detailed Description of the "test\_cv" Setting}}\label{detailed-description-of-the-test_cv-setting}}

It uses the loss function specfied in \texttt{fun\_control} and the
metric specified in \texttt{fun\_control}.

\begin{enumerate}
\def\labelenumi{\arabic{enumi}.}
\tightlist
\item
  First, the method \texttt{HyperTorch().fun\_torch} is called.
\item
  \texttt{fun\_torc()} calls
  \texttt{spotPython.torch.traintest.evaluate\_cv()} as follows:
\end{enumerate}

\begin{Shaded}
\begin{Highlighting}[]
\NormalTok{df\_eval, \_ = evaluate\_cv(}
\NormalTok{    model,}
\NormalTok{    dataset=fun\_control["test"],}
\NormalTok{    shuffle=self.fun\_control["shuffle"],}
\NormalTok{    device=self.fun\_control["device"],}
\NormalTok{    show\_batch\_interval=self.fun\_control["show\_batch\_interval"],}
\NormalTok{    task=self.fun\_control["task"],}
\NormalTok{    writer=self.fun\_control["writer"],}
\NormalTok{    writerId=config\_id,}
\NormalTok{)}
\end{Highlighting}
\end{Shaded}

Note: The data set \texttt{fun\_control{[}"test"{]}} is used for CV. The
rest is the same as for the \texttt{"train\_cv"} setting.

\hypertarget{sec-final-model-evaluation}{%
\subsubsection{Detailed Description of the Final Model Training and
Evaluation}\label{sec-final-model-evaluation}}

\hypertarget{detailed-description-of-the-train_tuned-procedure}{%
\paragraph{\texorpdfstring{Detailed Description of the
\texttt{"train\_tuned}
Procedure}{Detailed Description of the "train\_tuned Procedure}}\label{detailed-description-of-the-train_tuned-procedure}}

\texttt{train\_tuned()} is just a wrapper to
\texttt{evaluate\_hold\_out} using the \texttt{train} data set. It is
implemented as follows:

\begin{Shaded}
\begin{Highlighting}[]
\NormalTok{def train\_tuned(}
\NormalTok{    net,}
\NormalTok{    train\_dataset,}
\NormalTok{    shuffle,}
\NormalTok{    loss\_function,}
\NormalTok{    metric,}
\NormalTok{    device=None,}
\NormalTok{    show\_batch\_interval=10\_000,}
\NormalTok{    path=None,}
\NormalTok{    task=None,}
\NormalTok{    writer=None,}
\NormalTok{):}
\NormalTok{    evaluate\_hold\_out(}
\NormalTok{        net=net,}
\NormalTok{        train\_dataset=train\_dataset,}
\NormalTok{        shuffle=shuffle,}
\NormalTok{        test\_dataset=None,}
\NormalTok{        loss\_function=loss\_function,}
\NormalTok{        metric=metric,}
\NormalTok{        device=device,}
\NormalTok{        show\_batch\_interval=show\_batch\_interval,}
\NormalTok{        path=path,}
\NormalTok{        task=task,}
\NormalTok{        writer=writer,}
\NormalTok{    )}
\end{Highlighting}
\end{Shaded}

The \texttt{test\_tuned()} procedure is implemented as follows:

\begin{Shaded}
\begin{Highlighting}[]
\NormalTok{def test\_tuned(net, shuffle, test\_dataset=None, loss\_function=None,}
\NormalTok{    metric=None, device=None, path=None, task=None):}
\NormalTok{    batch\_size\_instance = net.batch\_size}
\NormalTok{    removed\_attributes, net = get\_removed\_attributes\_and\_base\_net(net)}
\NormalTok{    if path is not None:}
\NormalTok{        net.load\_state\_dict(torch.load(path))}
\NormalTok{        net.eval()}
\NormalTok{    try:}
\NormalTok{        device = getDevice(device=device)}
\NormalTok{        if torch.cuda.is\_available():}
\NormalTok{            device = "cuda:0"}
\NormalTok{            if torch.cuda.device\_count() \textgreater{} 1:}
\NormalTok{                print("We will use", torch.cuda.device\_count(), "GPUs!")}
\NormalTok{                net = nn.DataParallel(net)}
\NormalTok{        net.to(device)}
\NormalTok{        valloader = torch.utils.data.DataLoader(}
\NormalTok{            test\_dataset, batch\_size=int(batch\_size\_instance),}
\NormalTok{            shuffle=shuffle, }
\NormalTok{            num\_workers=0}
\NormalTok{        )}
\NormalTok{        metric\_value, loss = validate\_one\_epoch(}
\NormalTok{            net, valloader=valloader, loss\_function=loss\_function,}
\NormalTok{            metric=metric, device=device, task=task}
\NormalTok{        )}
\NormalTok{        df\_eval = loss}
\NormalTok{        df\_metric = metric\_value}
\NormalTok{        df\_preds = np.nan}
\NormalTok{    except Exception as err:}
\NormalTok{        print(f"Error in Net\_Core. Call to test\_tuned() failed. \{err=\}, }
\NormalTok{            \{type(err)=\}")}
\NormalTok{        df\_eval = np.nan}
\NormalTok{        df\_metric = np.nan}
\NormalTok{        df\_preds = np.nan}
\NormalTok{    add\_attributes(net, removed\_attributes)}
\NormalTok{    print(f"Final evaluation: Validation loss: \{df\_eval\}")}
\NormalTok{    print(f"Final evaluation: Validation metric: \{df\_metric\}")}
\NormalTok{    print("{-}{-}{-}{-}{-}{-}{-}{-}{-}{-}{-}{-}{-}{-}{-}{-}{-}{-}{-}{-}{-}{-}{-}{-}{-}{-}{-}{-}{-}{-}{-}{-}{-}{-}{-}{-}{-}{-}{-}{-}{-}{-}{-}{-}{-}{-}")}
\NormalTok{    return df\_eval, df\_preds, df\_metric}
\end{Highlighting}
\end{Shaded}

\newpage{}

\hypertarget{references}{%
\section*{References}\label{references}}
\addcontentsline{toc}{section}{References}

\hypertarget{refs}{}
\begin{CSLReferences}{1}{0}
\leavevmode\vadjust pre{\hypertarget{ref-bart21i}{}}%
Bartz, Eva, Thomas Bartz-Beielstein, Martin Zaefferer, and Olaf
Mersmann, eds. 2022. \emph{{Hyperparameter Tuning for Machine and Deep
Learning with R - A Practical Guide}}. Springer.

\leavevmode\vadjust pre{\hypertarget{ref-bart23e}{}}%
Bartz-Beielstein, Thomas. 2023. {``{PyTorch} Hyperparameter Tuning with
{SPOT}: Comparison with {Ray Tuner} and Default Hyperparameters on
{CIFAR10}.''}
\url{https://github.com/sequential-parameter-optimization/spotPython/blob/main/notebooks/14_spot_ray_hpt_torch_cifar10.ipynb}.

\leavevmode\vadjust pre{\hypertarget{ref-Bart13j}{}}%
Bartz-Beielstein, Thomas, Jürgen Branke, Jörn Mehnen, and Olaf Mersmann.
2014. {``Evolutionary Algorithms.''} \emph{Wiley Interdisciplinary
Reviews: Data Mining and Knowledge Discovery} 4 (3): 178--95.

\leavevmode\vadjust pre{\hypertarget{ref-bart20gArxiv}{}}%
Bartz-Beielstein, Thomas, Carola Doerr, Jakob Bossek, Sowmya
Chandrasekaran, Tome Eftimov, Andreas Fischbach, Pascal Kerschke, et al.
2020. {``Benchmarking in Optimization: Best Practice and Open Issues.''}
arXiv. \url{https://arxiv.org/abs/2007.03488}.

\leavevmode\vadjust pre{\hypertarget{ref-BLP05}{}}%
Bartz-Beielstein, Thomas, Christian Lasarczyk, and Mike Preuss. 2005.
{``{Sequential Parameter Optimization}.''} In \emph{{Proceedings 2005
Congress on Evolutionary Computation (CEC'05), Edinburgh, Scotland}},
edited by B McKay et al., 773--80. Piscataway NJ: {IEEE Press}.

\leavevmode\vadjust pre{\hypertarget{ref-Torczon00}{}}%
Lewis, R M, V Torczon, and M W Trosset. 2000. {``{Direct search methods:
Then and now}.''} \emph{Journal of Computational and Applied
Mathematics} 124 (1--2): 191--207.

\leavevmode\vadjust pre{\hypertarget{ref-Li16a}{}}%
Li, Lisha, Kevin Jamieson, Giulia DeSalvo, Afshin Rostamizadeh, and
Ameet Talwalkar. 2016. {``{Hyperband: A Novel Bandit-Based Approach to
Hyperparameter Optimization}.''} \emph{arXiv e-Prints}, March,
arXiv:1603.06560.

\leavevmode\vadjust pre{\hypertarget{ref-Meignan:2015vp}{}}%
Meignan, David, Sigrid Knust, Jean-Marc Frayet, Gilles Pesant, and
Nicolas Gaud. 2015. {``{A Review and Taxonomy of Interactive
Optimization Methods in Operations Research}.''} \emph{ACM Transactions
on Interactive Intelligent Systems}, September.

\leavevmode\vadjust pre{\hypertarget{ref-mont20a}{}}%
Montiel, Jacob, Max Halford, Saulo Martiello Mastelini, Geoffrey
Bolmier, Raphael Sourty, Robin Vaysse, Adil Zouitine, et al. 2021.
{``River: Machine Learning for Streaming Data in Python.''}

\leavevmode\vadjust pre{\hypertarget{ref-pyto23a}{}}%
PyTorch. 2023a. {``Hyperparameter Tuning with Ray Tune.''}
\url{https://pytorch.org/tutorials/beginner/hyperparameter_tuning_tutorial.html}.

\leavevmode\vadjust pre{\hypertarget{ref-pyto23b}{}}%
---------. 2023b. {``Training a Classifier.''}
\url{https://pytorch.org/tutorials/beginner/blitz/cifar10_tutorial.html}.

\end{CSLReferences}
\end{document}